\documentclass[lettersize,journal]{IEEEtran}
\usepackage{amsmath,amsfonts}
\usepackage{algorithmic}
\usepackage{algorithm}
\usepackage{array}
\usepackage{booktabs} 
\usepackage{makecell} 
\usepackage{textcomp}
\usepackage{stfloats}
\usepackage{url}
\usepackage{verbatim}
\usepackage{graphicx}
\usepackage{cite}
\usepackage[hidelinks, colorlinks]{hyperref}

\usepackage{listings}%
\usepackage{tabularx,tabulary, booktabs}
\usepackage{subcaption} 

\usepackage{color}
\usepackage[dvipsnames]{xcolor}
\usepackage{colortbl}
\usepackage{xspace}
\usepackage[utf8]{inputenc} 
\usepackage[T1]{fontenc}    

\usepackage{tcolorbox} 
\usepackage{multicol} 
\usepackage{xcolor}

\definecolor{Gray}{gray}{0.85}
\newcommand{\Gray}[0]{\rowcolor{gray!20}}
\newcommand{\Lgray}[0]{\rowcolor{gray!10}}

\definecolor{emerald}{rgb}{0.31, 0.78, 0.37}
\definecolor{coralred}{rgb}{1.0, 0.25, 0.25}

\begin{document}

\title{Enhancing Ultra High Resolution Remote Sensing Imagery Analysis with ImageRAG}

\author{Zilun Zhang $\dagger$, 
        Haozhan Shen $\dagger$,
        Tiancheng Zhao,
        Zian Guan, 
        Bin Chen, \\
        Yuhao Wang,
        Xu Jia,
        Yuxiang Cai,
        Yongheng Shang,
        Jianwei Yin

\thanks{$\dagger$: Equal Contribution}


\thanks{
Zilun Zhang, Haozhan Shen, Yuhao Wang, Yuxiang Cai, Yongheng Shang, and Jianwei Yin are with the College of Computer Science and Technology, Zhejiang University, Hangzhou, China; Tiancheng Zhao is with the Binjiang Research Institute of Zhejiang University; Bin Chen and Xu Jia are with the School of Software Engineering of Zhejiang University, Zian Guan is with the Polytechnic Institute of Zhejiang University. Corresponding author: Jianwei Yin. E-mail: zilun.zhang@zju.edu.cn; tianchez@zju-bj.com. 
}

}

\markboth{Journal of \LaTeX\ Class Files,~Vol.~14, No.~8, August~2021}%
{Shell \MakeLowercase{\textit{et al.}}: A Sample Article Using IEEEtran.cls for IEEE Journals}

\maketitle

\begin{abstract}
Ultra High Resolution (UHR) remote sensing imagery (RSI) (e.g. 10,000 $\times$ 10,000 pixels) poses a significant challenge for current Remote Sensing Vision Language Models (RSVLMs). If choose to resize the UHR image to standard input image size, the extensive spatial and contextual information that UHR images contain will be neglected. Otherwise, the original size of these images often exceeds the token limits of standard RSVLMs, making it difficult to process the entire image and capture long-range dependencies to answer the query based on the abundant visual context. In this paper, we introduce ImageRAG for RS, a framework to address the complexities of analyzing UHR remote sensing imagery with a little training requirement. By transforming UHR remote sensing image analysis task to image's long context selection task, we design an innovative image contextual retrieval mechanism based on the Retrieval-Augmented Generation (RAG) technique, denoted as ImageRAG.
ImageRAG's core innovation lies in its ability to selectively retrieve and focus on the most relevant portions of the UHR image as visual contexts that pertain to a given query. Fast path and slow path are proposed in this framework to handle this task efficiently and effectively. ImageRAG allows RSVLMs to manage extensive context and spatial information from UHR RSI, ensuring the analysis is both accurate and efficient. Code will be released in \url{https://github.com/om-ai-lab/ImageRAG}.

\end{abstract}

\section{Introduction}

In the field of remote sensing (RS), ultra-high-resolution (UHR) images often cover vast areas, encompassing diverse landscapes and a wide range of geospatial features. For deep learning applications such as visual question answering, semantic segmentation, object detection, and change detection, processing these large-scale images directly poses significant challenges. The high spatial resolution results in massive image sizes, making it difficult to directly train neural networks end-to-end with such images due to the hardware limitation. Additionally, the variability in scale, class distribution, and object sizes within these large images can lead to suboptimal performance if not handled properly. To address these, a common preprocessing step is to cut the original UHR images into smaller patches (e.g. $224 \times 224$ pixels or $512 \times 512$ pixels) \cite{dota} \cite{isaid} that can fit in regular deep learning workflows.

Multimodal Large Language Models (MLLM, in this paper we specifically refer to generative Vision-Language Models using Large Language Models as base model) such as Geochat \cite{geochat}, EarthGPT \cite{earthgpt}, SkysenseGPT \cite{skysensegpt}, VHM \cite{pang2024vhmversatilehonestvision}, etc. have demonstrated remarkable potential in RS tasks, including image captioning, visual grounding, relation reasoning, object detection, and visual question answering (VQA). 
However, the input image resolutions for these Remote Sensing Multimodal Large Language Models (RSMLLMs) are often limited and relatively small compared with the original satellite image. For example, models like LLaVA1.5 \cite{liu2023improvedllava} and VHM \cite{pang2024vhmversatilehonestvision} utilize image inputs of $336 \times 336$ pixels, while Geochat \cite{geochat} and SkysenseGPT \cite{skysensegpt} process images at $504 \times 504$ pixels. Contrastive Vision-language models (VLMs) specifically trained for RS, such as GeoRSCLIP \cite{rs5m} and RemoteCLIP \cite{remoteclip}, work with even smaller inputs, typically at $224 \times 224$ pixels.


\begin{figure}[t]
    \centering
    \includegraphics[width=0.49\textwidth]{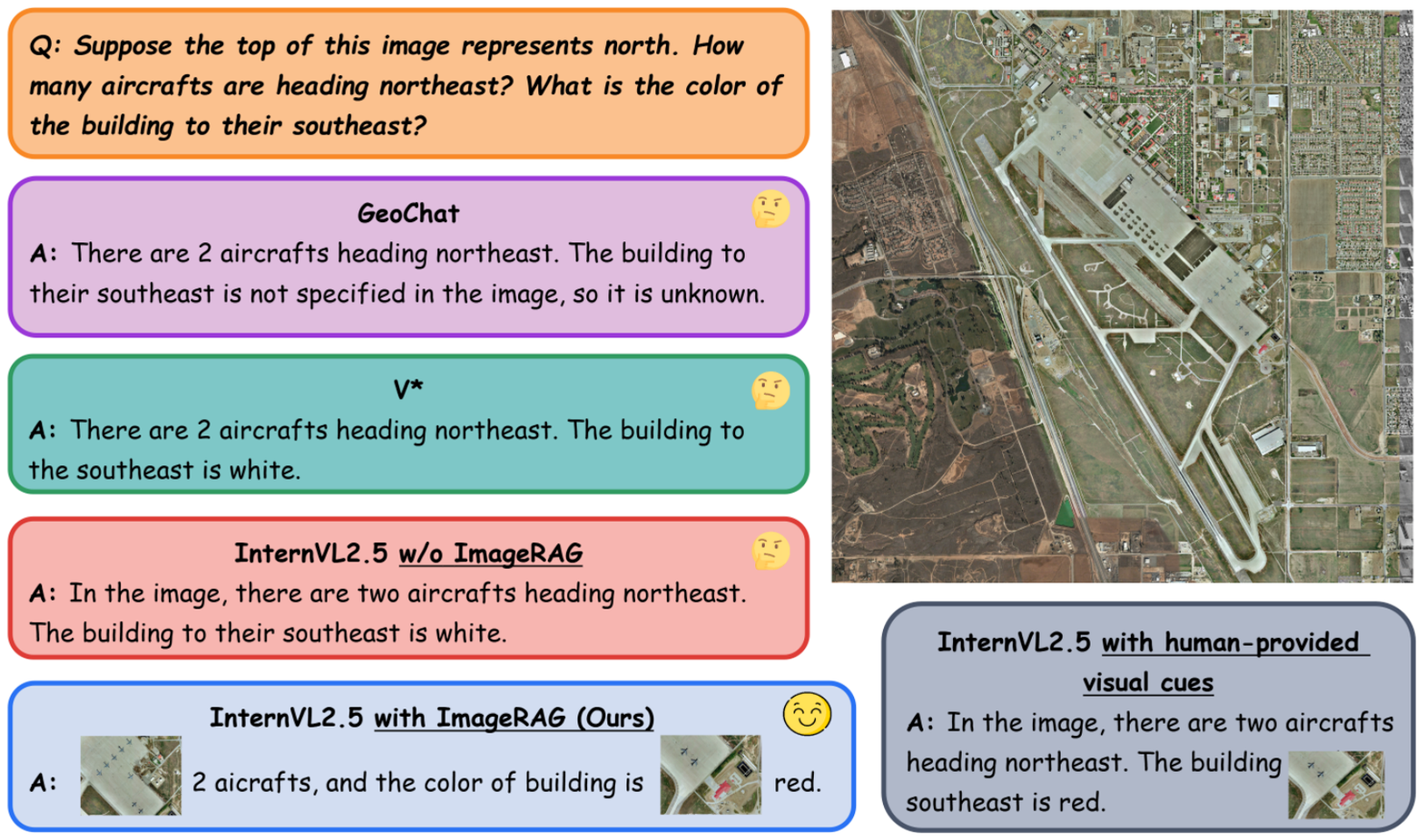}
    \caption{An example of challenging VQA task that requires analyzing small targets in a high-resolution image. Models such as GeoChat, InternVL2.5, and $V^{*}$ failed to answer. InternVL2.5 with the aid of ImageRAG and InternVL2.5 with human-provided visual cue can answer the question correctly.}
    \label{fig:teaser}
\end{figure}

\textcolor{black}{
Recently, Wang et al. proposed XLRS‑Bench \cite{wang2025xlrsbenchmultimodalllmsunderstand}, a benchmark for evaluating MLLMs on UHR RSI. Li et al. introduced STAR, a large‑scale dataset for scene‑graph generation in very‑high‑resolution satellite images, containing rich texture information (more than 400k <subject, relationship, object> triplets). Zhang et al. presented the MME‑RealWorld dataset \cite{mmerealworld}, which includes a remote‑sensing VQA subset with UHR images. These datasets could be potential benchmarks for VLMs based large remote sensing image understanding. Regarding specialized models, Luo et al. proposed a coarse‑to‑fine, text‑guided token‑pruning approach for large remote‑sensing images understanding \cite{luo2025largevisionlanguagemodelmeets}; however, its performance still leaves room for improvement.
}

In Figure \ref{fig:teaser}, we provide a qualitative example to illustrate how current RSMLLMs struggle to answer a challenging question that requires identifying small objects in a high-resolution image. The model's limitations in handling fine details and distinguishing small features become evident, leading to inaccurate responses when tasked with analyzing such intricate visual information (the model can answer correctly when the zoom-in image is provided).

To quantitatively identify this problem, we designed an experiment. We tested several RSMLLMs on a remote sensing subset of the MME-RealWorld benchmark (a VQA dataset with high-resolution images and tiny targets; details are provided in Section \ref{sec:MME-RealWorld-RS}) with 3738 questions. This subset is denoted as \textbf{MME-RealWorld-RS} in our paper. \textcolor{black}{The evaluation metric is the overall accuracy of regular VQA task (see section \ref{sec:rvqat} for detail).} RSMLLMs include InternVL2.5 (input image resolution: $448 \times 448$ pixels) \cite{internvl25}, SkysenseGPT (input image resolution: $504 \times 504$ pixels) \cite{skysensegpt}, Geochat (input image resolution: $504 \times 504$ pixels) \cite{geochat}, and VHM (input image resolution: $336 \times 336$ pixels) \cite{pang2024vhmversatilehonestvision}. The score for LLaVA1.5 (input image resolution: $336 \times 336$ pixels) \cite{liu2023improvedllava} is listed as a baseline, since the model structures of Geochat, SkysenseGPT, and VHM are derived from LLaVA1.5. The input image resolutions for these models are listed, and the input image will be resized to such fixed resolutions during training. Dynamic Resolution (DR) technique from InternVL 1.5 \cite{internvl15} is an input image preprocessing approach that divides the images into $448 \times 448$ pixel tiles based on the aspect ratio and resolution of the input images. An increased number of tiles allows for a higher degree of image magnification, thereby enabling the observation of finer details within the image. InternVL2.5 using Dynamic Resolution with a max number of 6 and 12 dynamic tiles are compared as well (InternVL + DR6 and InternVL + DR12 in the figure).

\begin{figure}[t]
    \centering
    \includegraphics[width=0.49\textwidth]{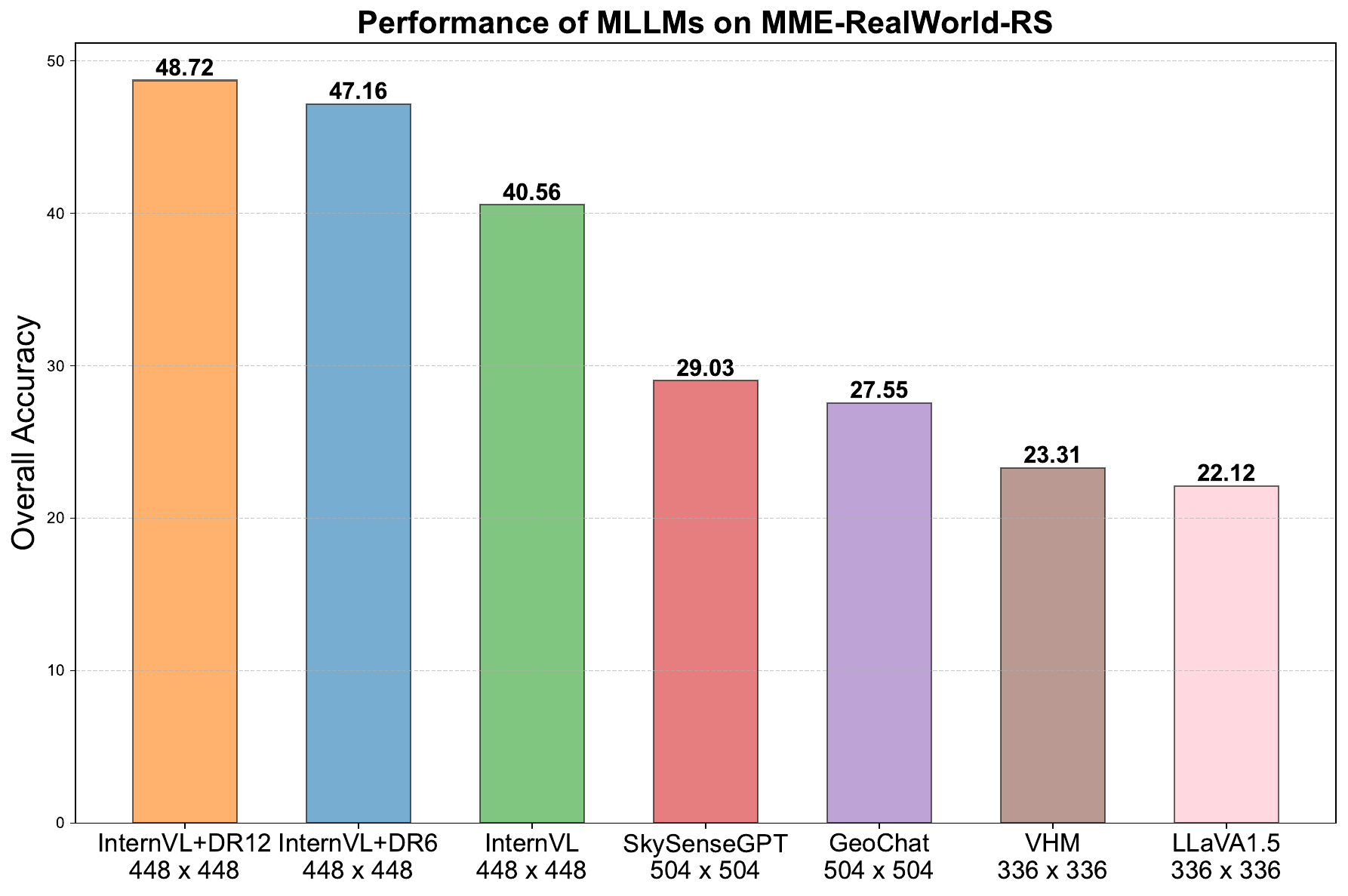}
    \caption{The performance of MLLMs in the MME-RealWorld-RS benchmark's remote sensing subset. The specified image resolutions for model input are listed in the end. “DR” represents the \textbf{Dynamic Resolution} technique, with 6 or 12 indicating the maximum number of tiles obtained through Dynamic Resolution. In general, model performance tends to improve with increased input image resolution (DR can be seen as an enhancement of input image resolution)}
    \label{fig:mllms_mmerealworld}
\end{figure}

Figure \ref{fig:mllms_mmerealworld} presents the results of this experiment. \textbf{In general, model performance tends to improve with increased input image resolution} (the Dynamic Resolution technique can be seen as an enhancement of input image resolution, and the level of enhancement grows as the maximum number of dynamic tiles increases). That makes sense because increasing the input image resolution magnifies tiny objects, allowing for better detection and analysis of small details in UHR RSI. InternVL2.5 shows better performance than SkysenseGPT and GeoChat, even with a smaller input image size. This could be attributed to the fact that SkysenseGPT and GeoChat utilize a vision encoder that is pre-trained with an input resolution of $336 \times 336$ pixels and interpolate the input image resolution to $504 \times 504$ pixels during the Supervised Fine-tuning stage. In contrast, the vision encoder of InternVL2.5 is trained from scratch with an input image resolution of $448 \times 448$ pixels.

The trend of this experiment can be interpreted in another way: If models resize the input image to a much lower resolution (compared with the original image size), image details such as tiny objects will become hard to notice and may be neglected when the model thinks and generates answers. This makes MLLMs difficult to apply in UHR RSI. \textcolor{black}{We identify four types of approaches for applying MLLMs to UHR RSI, each with its own set of limitations.} 

\textcolor{black}{
The first approach involves resizing UHR images to a smaller size in order to be compatible with current MLLMs. Most RSMLLMs use a fixed input image resolution and number of visual tokens because they initialize the model weights with a pretrained MLLM. Take Encoder-MLP-LLM structured RSMLLMs as an example (e.g., LLaVA-based models such as Geochat, VHM, and SkysenseGPT), they first resize images to 336 $\times$ 336, then divide the images into 576 visual patches of 14 $\times$ 14. Next, these visual patches are projected into 576 visual tokens. These tokens share a unified representation space with language tokens. Finally, they process the tokens using an LLM.
However, these operations significantly reduces the visibility of small objects in the images, making them challenging to detect, even for humans. For instance, VHM \cite{pang2024vhmversatilehonestvision} claims its difficulty in handling small objects, likely due to limitations in input image resolution of $336 \times 336$.
}

\textcolor{black}{
The second approach divides UHR images into smaller sub-images that can be sequentially processed by MLLMs. While this allows for compatibility with existing model architectures, it results in the loss of global and relative information and relationships present in the original large-scale image, as only portions of the image are considered at a time. 
}

\textcolor{black}{
The third approach references techniques from general LLMs for managing long context, such as Positional Interpolation \cite{chen2023extendingcontextwindowlarge} and LongROPE \cite{ding2024longropeextendingllmcontext}. Or adopting architecture from video MLLMs like LongVILA \cite{xue2024longvilascalinglongcontextvisual}, which can extend the context window effectively. These approach could enable the integration of entire UHR images while maintaining global information. However, this would require retraining the models from scratch and could be limited by LLM's context length. 
}

\textcolor{black}{
The fourth approach employs guided visual search methods that focus on relevant patches, such as $V^{*}$ \cite{wu2023vguidedvisualsearch}, or hybrid architectures like LongLLaVA \cite{wang2024longllavascalingmultimodalllms}, which enable the processing of very large input images, and not neglect the small targets. Similar to the drawbacks of the third approach, this method also requires retraining the model and demands task-specific annotations, adding to the complexity and effort needed.
}

Three crucial aspects for MLLMs to effectively handle UHR RSI are: (1) managing small targets, ensuring that the model can accurately aware and analyze fine details within images; (2) processing the UHR image in a way that integrates with MLLMs without significantly increasing the number of image tokens, which would lead to high computational costs; and (3) achieving these goals while minimizing the need for additional training or specialized annotation.

To address these problems, we contribute the \textbf{ImageRAG} framework, which offers several key advantages. 
\begin{itemize}
    \item It retrieves and emphasizes relevant visual context from the UHR image based on the text query, allowing the MLLM to focus on \textbf{important details, even tiny ones}.
    \item It integrates various \textbf{external knowledge sources} to guide the model, enhancing the understanding of the query and the UHR RSI.
    \item ImageRAG only requires a little amount of training, making it a practical solution for efficiently handling UHR RSI.
\end{itemize}


\section{Benchmark}
\subsection{MME-RealWorld-RS}
\label{sec:MME-RealWorld-RS}

\begin{figure*}[ht]
    \centering
    \begin{subfigure}[ht]{0.42\textwidth}
        \centering
        \includegraphics[width=\textwidth]{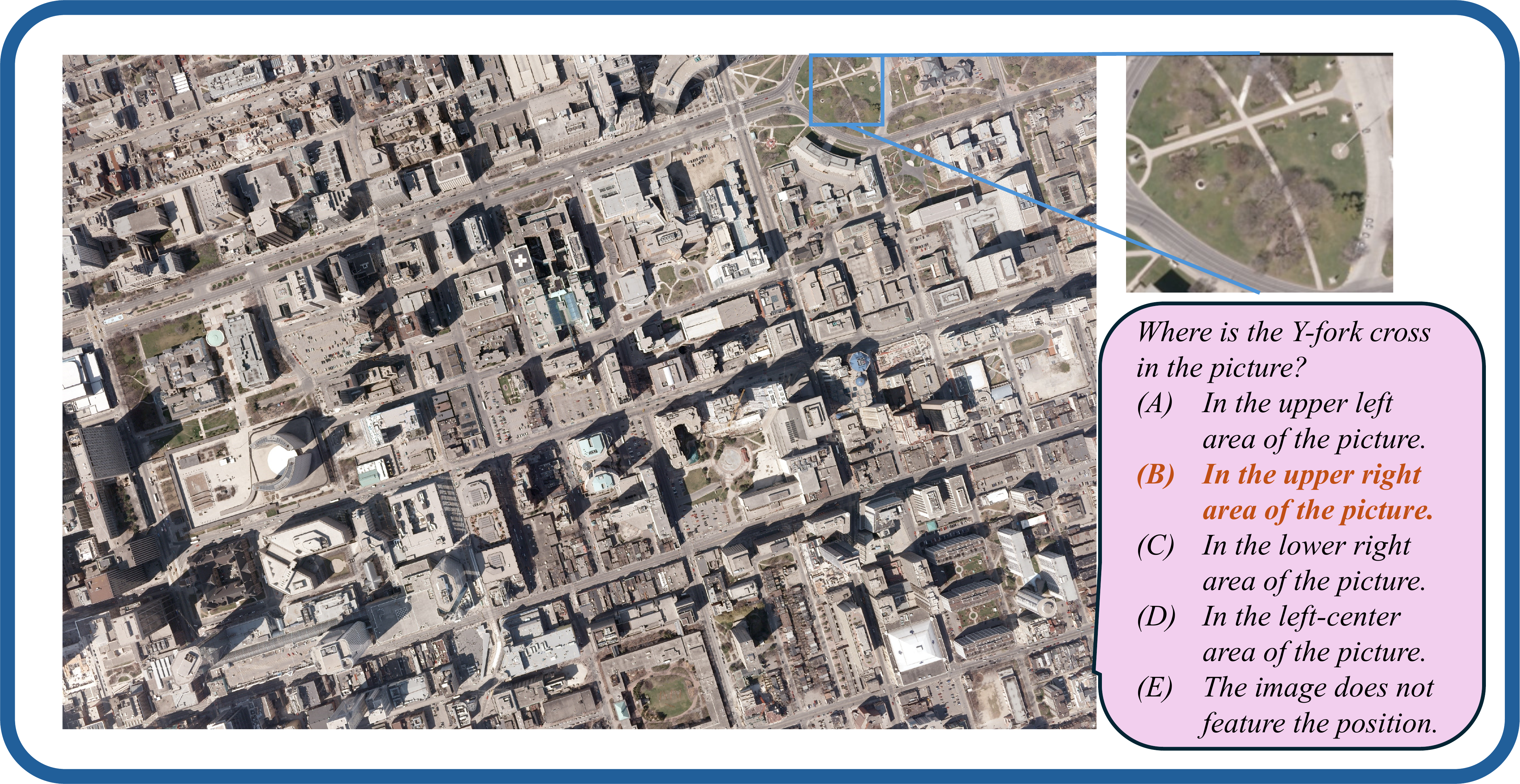}
        \caption{Spatial Relationship Task}
        \label{fig:sub1}
    \end{subfigure}
    \hfill
    \begin{subfigure}[ht]{0.45\textwidth}
        \centering
        \includegraphics[width=\textwidth]{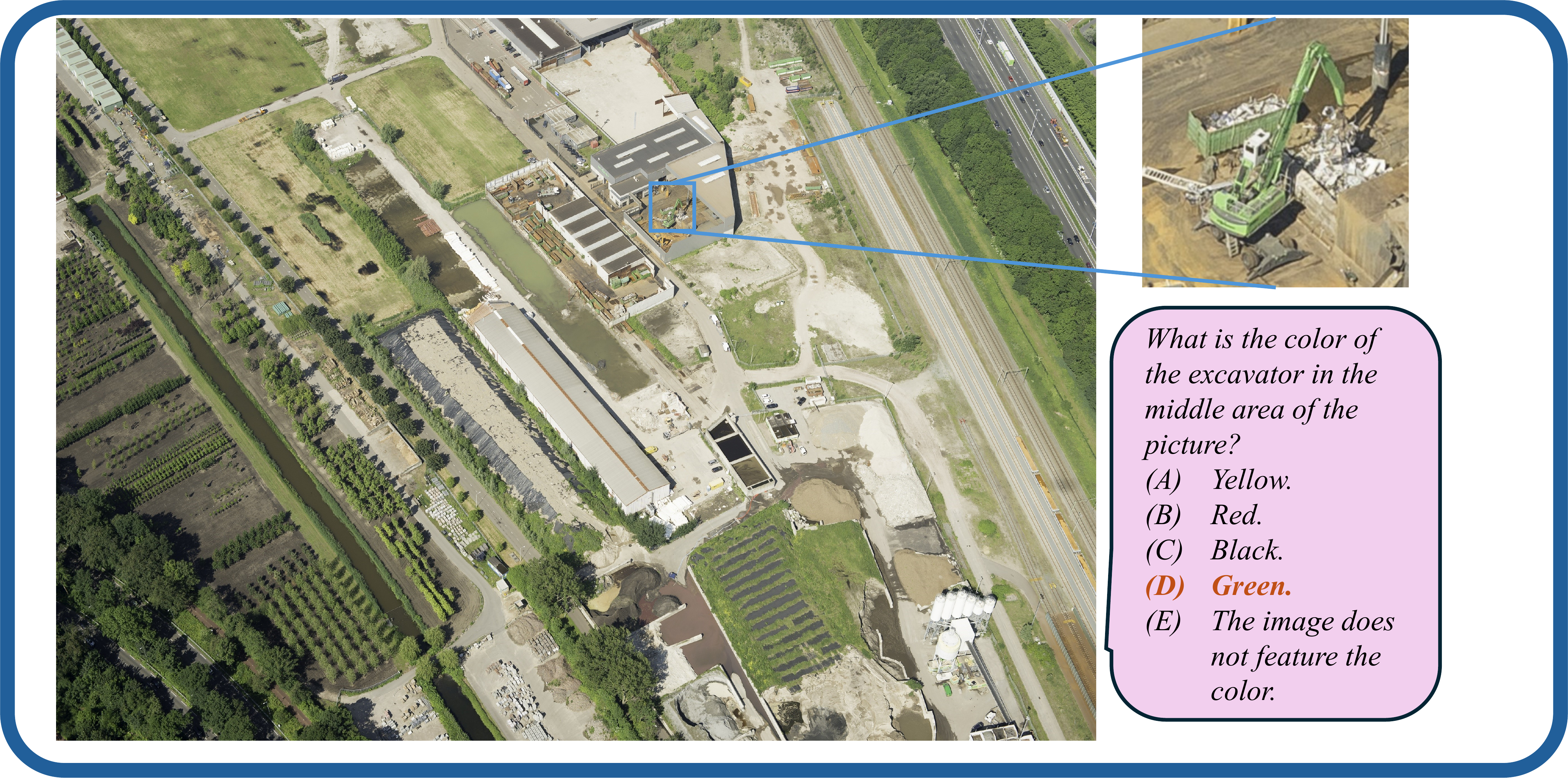}
        \caption{Color Recognition Task }
        \label{fig:sub2}
    \end{subfigure}

    \begin{subfigure}[ht]{0.42\textwidth}
        \centering
        \includegraphics[width=\textwidth]{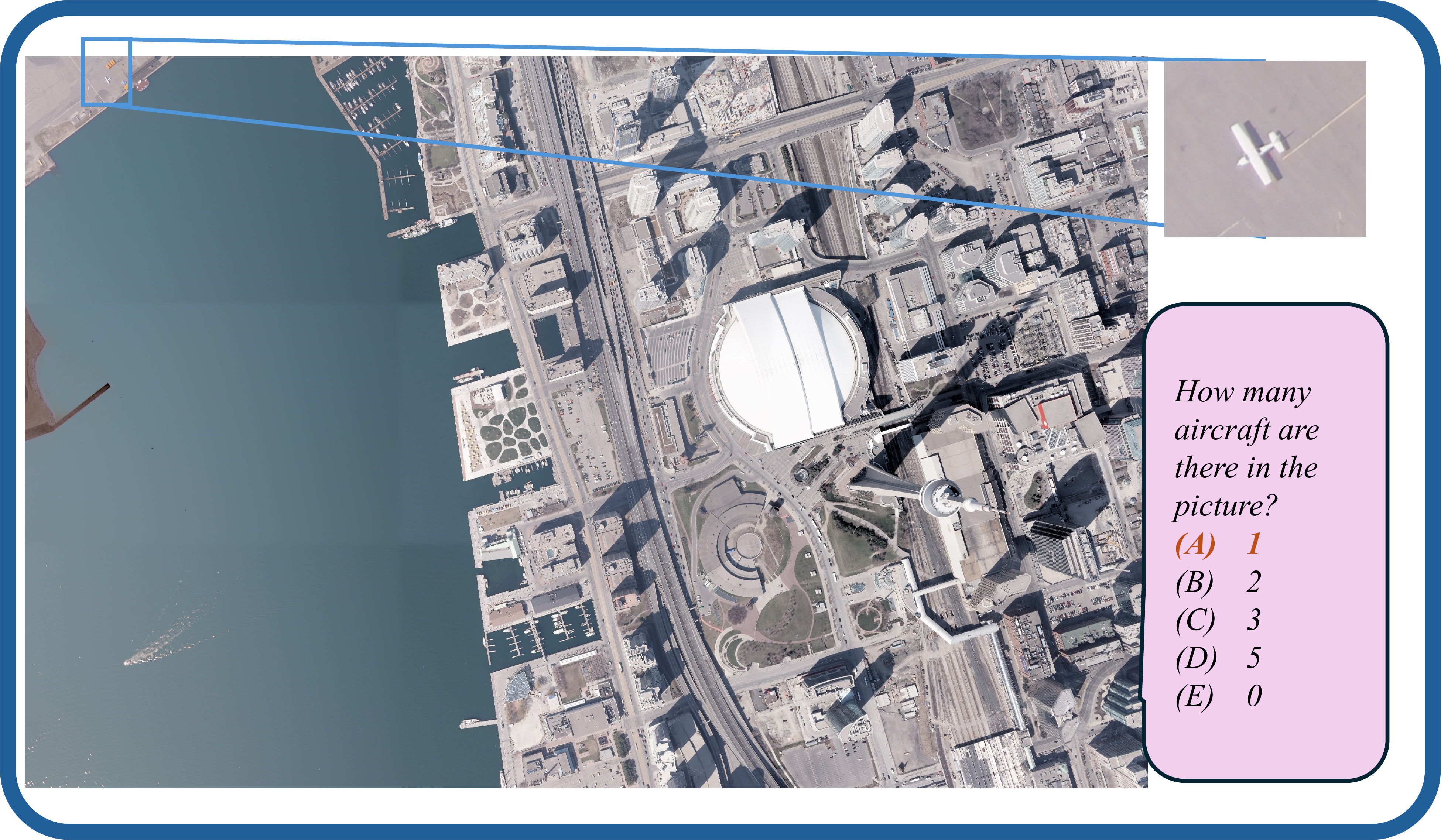}
        \caption{Object Counting Task}
        \label{fig:sub3}
    \end{subfigure}
    \hfill
    \begin{subfigure}[ht]{0.45\textwidth}
        \centering
        \includegraphics[width=\textwidth]{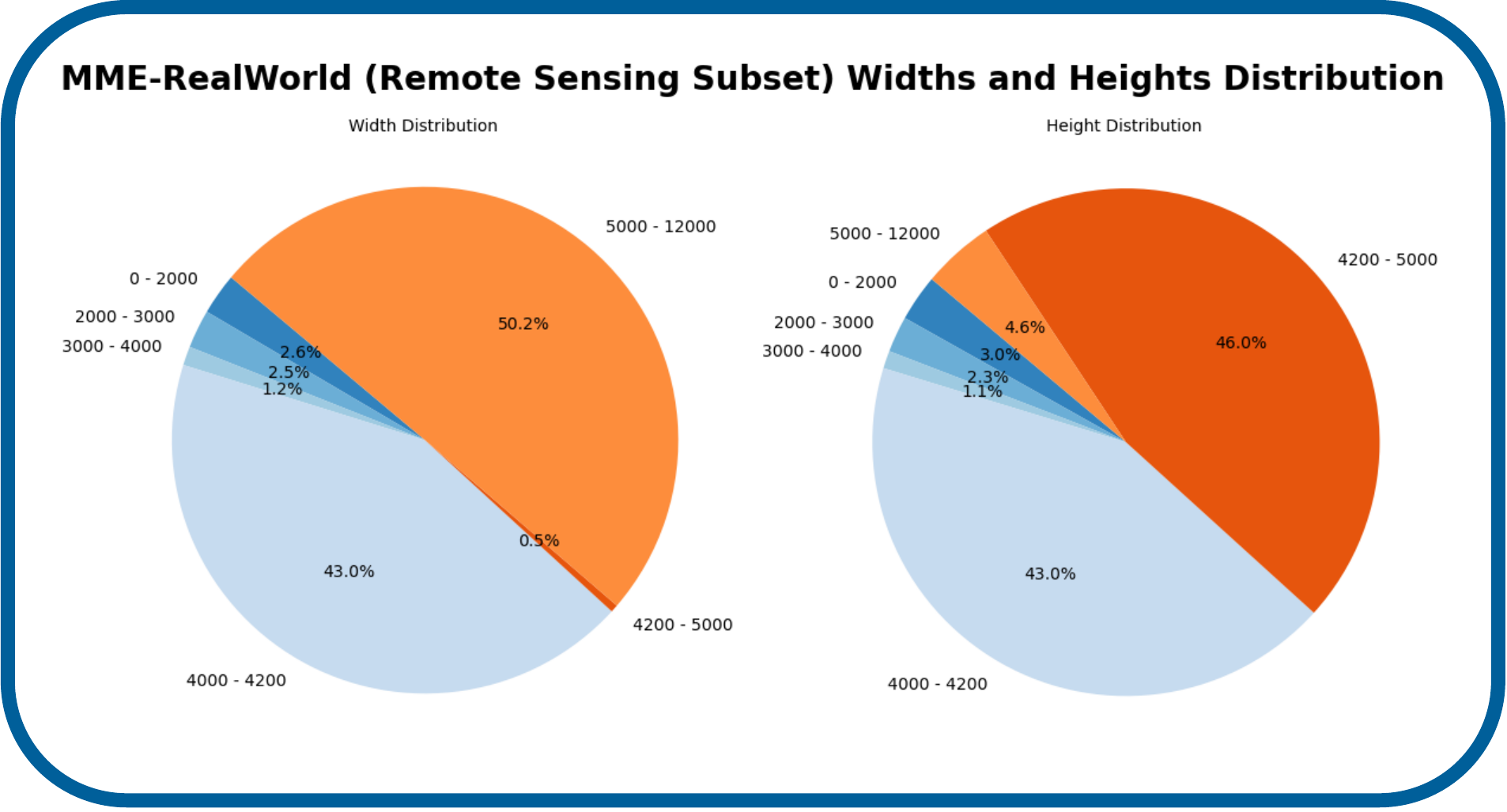}
        \caption{Image Size Statistics of MME-RealWorld-RS}
        \label{fig:sub4}
    \end{subfigure}

    \caption{Visualization of Subtasks from MME-RealWorld-RS Dataset and Statistics of the Images. Image examples are taken from the Appendix of MME-RealWorld \cite{mmerealworld} Paper.}
    \label{fig:demo_mmerealworld}
\end{figure*}

\textbf{The Remote Sensing subset of the MME-RealWorld benchmark} is designed to evaluate the capabilities of MLLMs in handling high-resolution images with VQAs (multiple choice questions). These images are characterized by their extremely high quality and rich details, making them challenging even for human annotators. The data sources include the FAIR1M dataset \cite{sun2021fair1mbenchmarkdatasetfinegrained}, the ISPRS Potsdam dataset \footnote{https://paperswithcode.com/dataset/isprs-potsdam}, VGoogle \cite{8746238}, VBing \cite{8746238}, and VArcGIS \cite{8746238}. These datasets are sourced from Google Earth, Bing World Imagery, and ArcGIS World Imagery. It includes sub-tasks such as spatial relationship understanding, color recognition, and object counting.

\textbf{Spatial Relationship Understanding} task (1,257 QA pairs) involves understanding the absolute and relative spatial relationships between objects in the images. \textbf{Color Recognition} task (1,226 QA pairs) requires the model to identify and describe the colors of specific objects in the images. \textbf{Object Counting} task (1,255 QA pairs) involves counting specific objects within the images, such as aircraft, vehicles, or buildings. These tasks are challenging due to the large image size and small target size, which can be easily overlooked. 

A demonstration of MME-RealWorld-RS is presented in Figure \ref{fig:demo_mmerealworld}, which includes examples of 3 subtasks and dataset statistics. This dataset comprises 1,265 high-resolution images that were manually selected from over 70,000 public remote sensing images. The average image size is $5,602 \times 4,445$ pixels across all related images (some images may be repeated in different tasks). There are 3,738 QA pairs in the dataset and the questions and answers are generated by 20 volunteers manually, while an expert examined the quality of the questions to ensure they conformed to the required standards. 

\subsection{MME-RealWorld-Lite-RS}
\label{sec:MME-RealWorld-Lite-RS}

The MME-RealWorld-RS dataset is an ideal benchmark for evaluating the ImageRAG framework due to its distinctive characteristics. It includes UHR RSI and features extremely tiny objects, challenging the framework's ability to detect and classify small-scale targets effectively and efficiently. In addition, it contains non-ordinary object classes, offering a diverse set of objects beyond the typical categories found in standard remote sensing benchmarks. This diversity ensures a comprehensive evaluation of the framework's capabilities. Lastly, target-of-interest in the question is unique (for Color Recognition task and Spatial Relationship Understanding task), simplifying the evaluation process and ensuring clear, unambiguous answer. These features collectively make MME-RealWorld-RS a robust and suitable benchmark for assessing the ImageRAG framework's performance. 


However, evaluating only the final accuracy for VQA questions is too simplistic. Similar to the RAG framework, we aim to determine whether ImageRAG can retrieve the correct and useful visual cues (e.g. image patches) to assist the MLLM in analyzing and determining the final answer. This requires the annotation of regions of interest (e.g., 2D coordinates), a feature that MME-RealWorld-RS does not inherently provide. 


To address this issue, we annotated MME-RealWorld-lite-RS, which is a RS subset of MME-RealWorld-lite \footnote{https://huggingface.co/datasets/yifanzhang114/MME-RealWorld-Lite }. Similar to MME-RealWorld-RS, it contains 150 QA pairs, with 50 for each of the three subtasks. \textbf{Specifically, we examined the VQA triplets one-by-one and labeled the coordinates of the rectangular Region-of-Interest (ROI) in the image, based on the provided question and answer.} The coordinates are in $[x_1,y_1,x_2,y_2]$, representing the x-y coordinates of the top-left and the bottom-right points of the ROI box. One annotator annotated the box, and two annotators examined the correctness of the region-of-interest for each VQA triplet. We established the following guidelines to guide our labeling process:
\begin{itemize}
\item One Region-of-Interest box per VQA triplet.
\item The Region-of-Interest box must contain all objects mentioned in the question.
\item The Region-of-Interest box should be as small as possible while still satisfying the previous condition.
\item Two annotators must have an agreement on the annotated Region-of-Interest box, and the uniqueness of target-of-interest must be checked.
\end{itemize}
We used the jupyter bbox widget \footnote{https://github.com/gereleth/jupyter-bbox-widget} to assist the label process. During the annotation process, we found that some labels from MME-RealWorld were incorrect or ambiguous. We corrected them and provided a correction list in the Appendix \ref{sec:appendix_correction}.


\begin{figure}[htbp]
    \centering
    \includegraphics[width=0.46\textwidth]{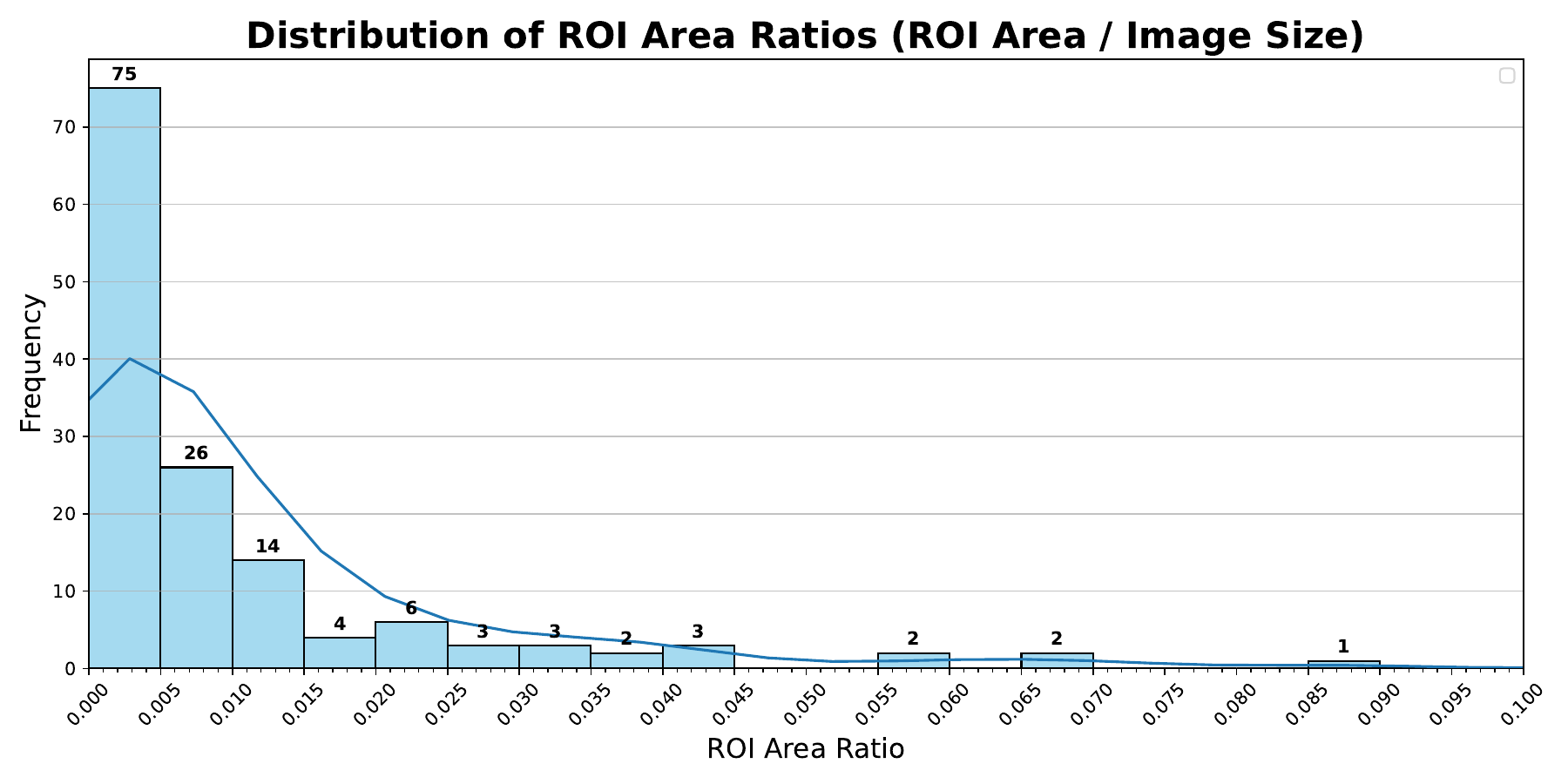}
    \caption{Distribution of ROI area ratios (ROI Area / Image Size)}
    \label{fig:ROI_area_dist}
\end{figure}

\begin{figure}[htbp]
    \centering
    \includegraphics[width=0.46\textwidth]{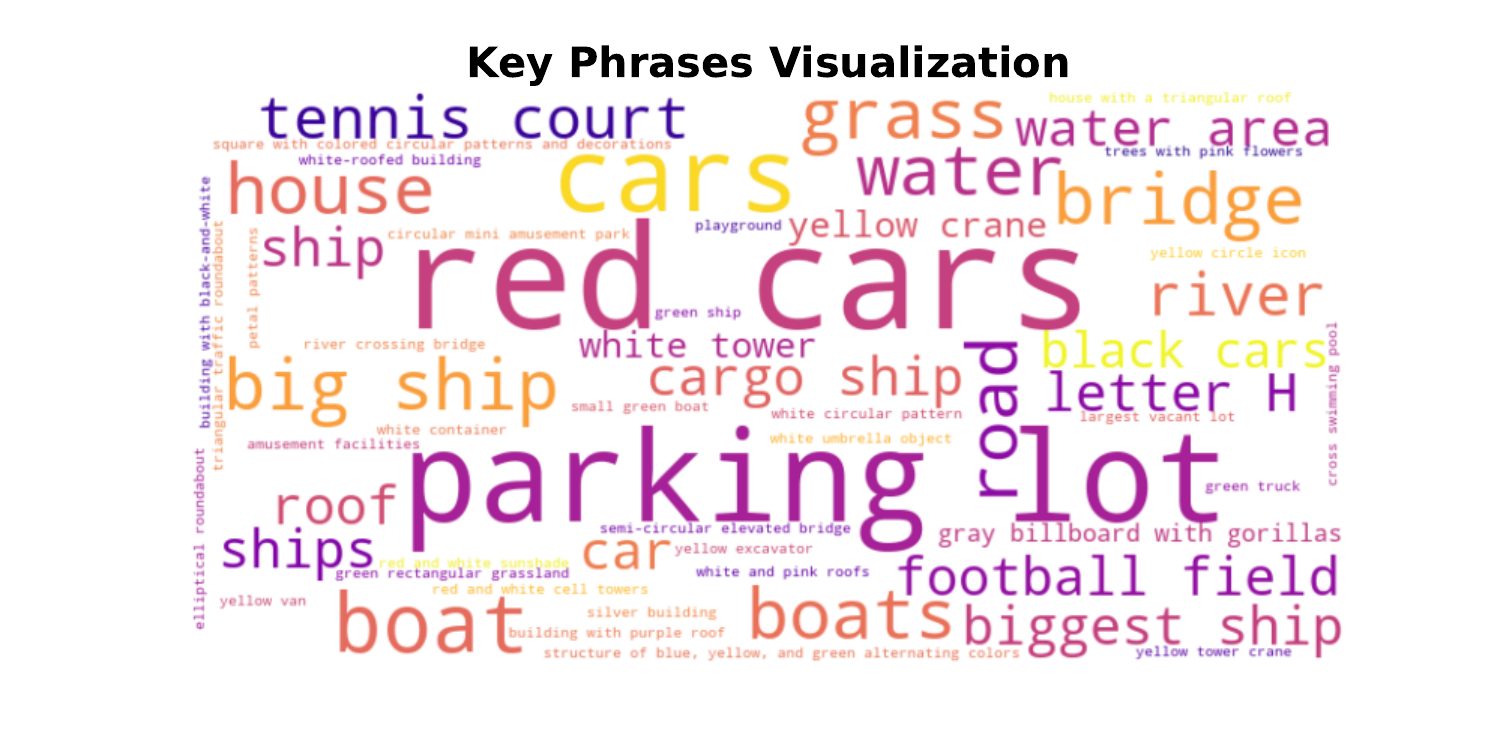}
    \caption{Key Phrases Word Cloud}
    \label{fig:keyphrases_mmerealworldlite_rs}
\end{figure}

Figure \ref{fig:ROI_area_dist} presents the distribution of ROI area ratios (ROI Area \/ Image Size) of MME-RealWorld-Lite-RS. The median of ROI area ratios is 0.00497, which means the ROI area usually takes up only a tiny portion of the entire UHR RSI. The extracted key phrases (from questions in MME-RealWorld-Lite-RS dataset) can be found in Figure \ref{fig:keyphrases_mmerealworldlite_rs}.

\section{Task}
\label{sec:task}
When evaluating ImageRAG using a VQA dataset with multiple-choice questions, \textbf{accuracy} is the most commonly used metric. However, this metric only \textbf{indirectly} reflects the capability of the ImageRAG model, as it does not directly assess the retrieval results. Accuracy can improve if the retrieved visual cues are closely related to the question and the MLLM has the capability to understand and utilize these visual cues to answer the question. Based on this analysis, we break down the assessment into two key components: (1) evaluating how closely the retrieved visual cues align with the ground truth region-of-interest box, and (2) assessing the ability of MLLMs to make judgments by inferring visual cues. In summary, we have three subtasks to evaluate the performance of ImageRAG: Visual Cue Retrieval Task, Inferring VQA Task, and Regular VQA task. We will expand former two in next subsections.

\subsection{Regular VQA Task}
\label{sec:rvqat}
The regular VQA task requires models to understand image content and provide accurate answers to given natural language questions. In the context of MME-RealWorld-RS and MME-RealWorld-lite-RS, models must respond with only a letter (A, B, C, D, or E) corresponding to the input question and image. Given a \text{MLLM}, a question $T_i$ and image $I_i$, the regular VQA task can be represented as follow ($R_i$ is the model output such as $A, B, C, D, E$):

\begin{align}
    R_i = \text{MLLM}(I_i, T_i)
\end{align}

The accuracy (\text{Acc}) in the context of the regular VQA task can be calculated using the following formula:

\begin{align}
    \text{Acc} = \frac{\text{\# of Questions with Correct Response}}{\text{\# of All Questions}}
\end{align}

\subsection{Inferring VQA Task}

For ordinary RAG framework, the retrieved context can be directly organized by a text prompt and fed into the LLM since they are both in text modality. The generalization ability of the LLM will automatically make the final decision by inferring the selected text context as evidence. However, this is non-trivial for MLLMs, which need to infer multiple visual contexts (as model input) from image modality and facilitate decision-making. Usually, MLLMs are not trained for such a target. To assess the capability of MLLMs in using visual cues to aid the final decision-making process, we propose the \textbf{Inferring VQA Task} for MLLMs.

The inferring VQA task, as hinted by the name, is a variant of the VQA task that involves providing visual context and prompt to infer. The task can be formalized as follow:

\begin{align}
    R_i = \text{MLLM}(I_i, V_{i}, T_i \mid \text{Prompt})
    \label{eq: MMLM_cues}
\end{align}

In this paper, the $V_{i}$ is the ROI box, which is an absolutely corrected visual cue. Same as VQA task, Overall Accuracy (of multiple choice questions) is the metric to evaluate the Inferring VQA Task. MLLM that achieves higher accuracy with given visual cues is a more effective inferring model. \textbf{Inferring model using ROI box can be considered as an upper bound of ImageRAG framework}. The evaluation metric of inferring VQA task is accuracy. This task is designed to evaluate the generation stage of ImageRAG framework.

\subsection{Visual Cue Retrieval Task}
Similar to the RAG framework, the goal of the \textbf{Visual Cue Retrieval Task} in ImageRAG is to retrieve useful image patches as visual cues (evidence) to assist MLLMs (inferring model) in making accurate decisions. Ideally, the retrieved visual cues should overlap with the ground truth region-of-interest box. Therefore, recall@k is chosen as the evaluation metric of this retrieval subtask.

For each question from the MME-RealWorld-Lite-RS dataset, the ImageRAG framework outputs at most $k$ visual cues $\{V_i\}_{i=1}^{k}$, and there is a ground truth ROI box ($G$) for each question, as we mentioned in section \ref{sec:MME-RealWorld-Lite-RS}. We define the \textbf{Recall@k with IoU threshold T} as follow:

\begin{align}
     \text{Recall@k} = \frac{1}{k} \sum_{i=1}^{k} \mathbb{I}\left( \mathbf{IoU}(V_i, G) \geq T \right)
\end{align}

Where $\mathbf{IoU}(V_i, G)$ is the regular intersection over union score between a visual cue $V_i$ and the ROI box $G$, which are boxes represented by 4-coordinate $[x_1, y_1, x_2, y_2]$, and $\mathbb{I}$ is the indicator function \footnote{https://en.wikipedia.org/wiki/Indicator\_function} depending on the IoU score and IoU threshold T. 

\begin{align}
\label{eq: indicator}
\mathbb{I}\left( \mathbf{IoU(V_i, G)} \geq T \right) = 
\begin{cases}
1, & \text{if } \mathbf{IoU}(V_i, G) \geq T \\
0, & \text{otherwise}
\end{cases}
\end{align}

Finally, the mean recall is calculated by taking the average of all Recall@k values across all questions in the MME-RealWorld-Lite-RS dataset. Mathematically, it can be represented as:

\begin{align}
     \label{eq: mr}
    \text{MR} = \frac{1}{N} \sum_{n=1}^{N} \text{Recall@k}_n
\end{align}

where N is the total number of questions in the dataset, and Recall@$k_n$ is the Recall@k with IoU threshold T calculated with the retrieved visual cues and the ROI box of n-th question. This task is designed to evaluate the retrieval stage of ImageRAG framework.

\section{Overview of The ImageRAG Framework}
\label{sec:imagerag}

\begin{figure}[ht!]
    \centering
    \includegraphics[width=0.45\textwidth]{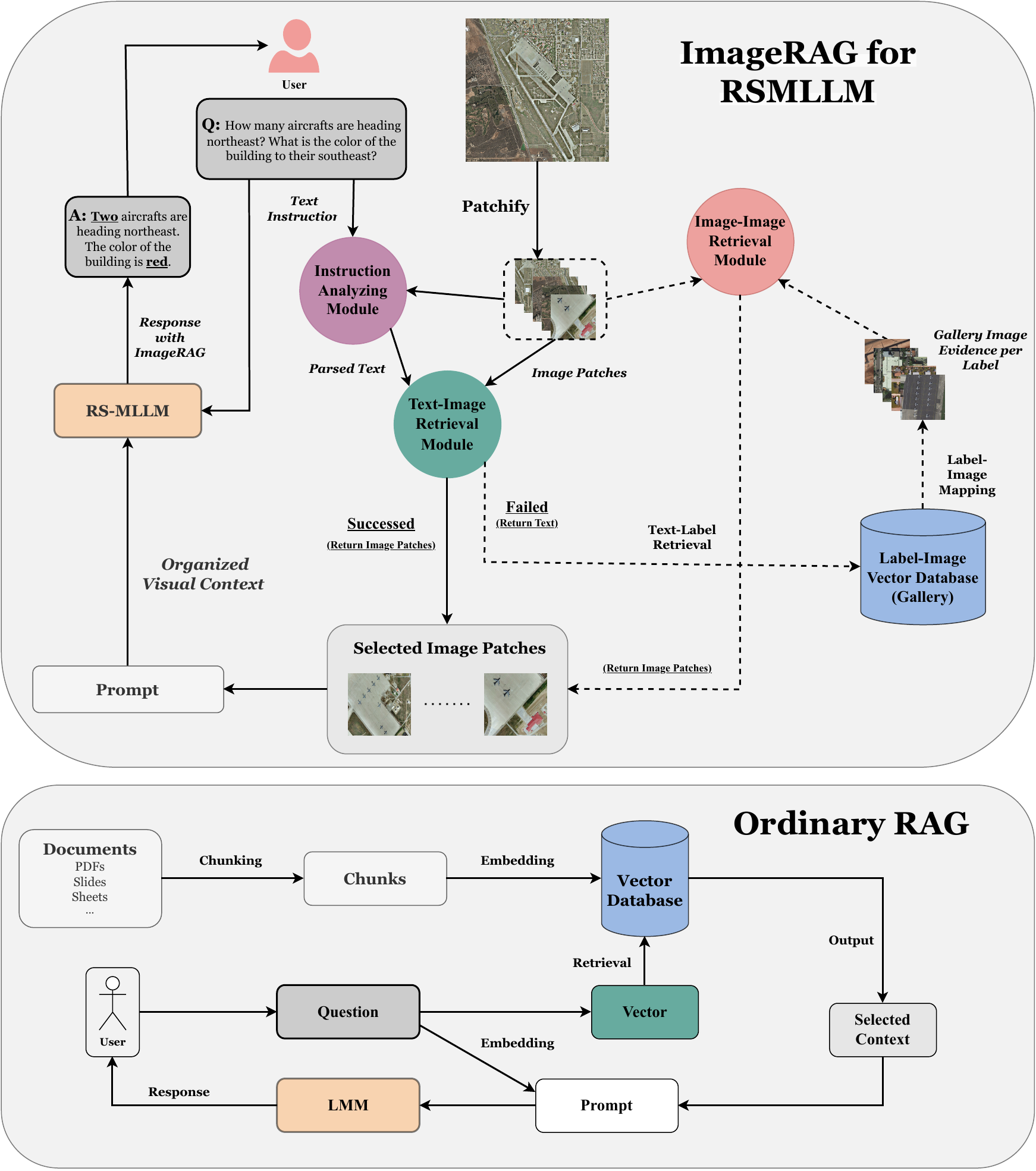}
    \caption{The ImageRAG Framework (up) and Ordinary RAG (down). \textbf{Dashed lines} represent the slow path of ImageRAG, and the \textbf{solid lines} represents the fast path. Detail introduction on ImageRAG can be found in section \ref{sec:imagerag}.}
    \label{fig:imagerag}
\end{figure}

As shown in Figure \ref{fig:imagerag} (down), ordinary RAG boosts the capabilities of LLMs by retrieving and referencing relevant document chunks from
external knowledge database through the semantic similarity. The process involves two main stages: Retrieval and Generation. In this way, RAG effectively reduces the problem such as generating factually incorrect and out-of-date content \cite{gao2024retrievalaugmentedgenerationlargelanguage}. Moreover, a domain-specialized LLM can be achieved by equipping the LLM with RAG and a domain knowledge database \cite{barron2024domainspecificretrievalaugmentedgenerationusing}.


A challenge in applying RAG to UHR RSI is how to extend RAG to visual modality. It requires VLMs to associate text and visual embeddings, which may face difficulties in aligning visual concepts in satellite views with corresponding text descriptions. Besides, consider image patches as contexts could result in no visual contexts be found to aid generation. We will expand this and provide a solution in section \ref{sec:imageragmotivation}.

Our ImageRAG framework for RSMLLM references the idea of RAG, but focusing on retrieving \textbf{visual contexts} as evidences for the text query.  As shown in Figure \ref{fig:imagerag} (up), ImageRAG also contains two stages, \textbf{Retrieval Stage} and \textbf{Generation Stage}. 
We denote a given image as $I_i$ and corresponding text query (instruction) as $T_i$. The ImageRAG framework aims to \textbf{retrieve a set of relevant visual contexts $V_{i}$}
to augment input and \textbf{generate response $R_i$}. There are two modes for ImageRAG to work, \textbf{fast path} and \textbf{slow path}. 

\subsection{Retrieval Stage}
\label{sec:retrievalstage}

Given an Image $I$, a text query $T$, a \textbf{Patch Division Algorithm} $F$, an \textbf{Question Analyzing Module} $G$, a \textbf{Text-Image Retrieval Module} $M_{\text{text-img}}$ (including an image encoder $f_{\text{img}}$, a text encoder $f_{\text{text}}$, a patch select function $H_{\text{fast}}$), a \textbf{Text-Image Vector Database} $D$ with threshold $\delta$, and an \textbf{Image-Image Retrieval Module} $M_{\text{img-img}}$ (including image encoder $f_{\text{img}}$ and patch select function $H_{\text{slow}}$). A set of visual cues $V$ can be selected using the following strategy:

\[
V = 
\begin{cases} 

M_{\text{text-img}}(I, T \mid (F, G, f_{\text{img}}, f_{\text{text}}, H_{\text{fast}})) \ \ \text{fast path}\\
M_{\text{img-img}}(I, T, D \mid (F, G, f_{\text{img}}, f_{\text{text}}, H_{\text{slow}})) \ \  \text{slow path}
\end{cases}
\]

\subsubsection{Image Patch Division Approach} 

There are several image patchification approaches. For instance, Vision Transformer  \cite{dosovitskiy2021imageworth16x16words} divides image into fixed-size, non-overlapping grids with dimensions such as $32 \times 32$ or $16 \times 16$ pixels. The Swin Transformer \cite{liu2021swintransformerhierarchicalvision} uses a hierarchical architecture to partition the image at multiple scales, employing shifted windows to capture more diverse contextual information. DetailCLIP \cite{zhang2023injectingimagedetailsclips} introduced the "Complete Cover" method, which aims to partition the image that patches can cover objects of any scale. Suppose the patch division approach $F$ outputs $m$ image patches for each image, the patchification process can be formulated as:
\begin{align}
    P = F(I)=\{p_i\}_{i=1}^m
\end{align}
Where $P$ denotes a \textbf{set of image patches} for image $I$.

\subsubsection{Question Analyzing Module} The instruction analyzing module $G$ processes the input question $T$ and extracts a set of key phrases $Q$ consisting of $n$ key phrases. The whole process is entirely based on the input text instruction.

\begin{align}
   Q = G(T) = \{t_{i}\}_{i=1}^n 
\end{align}

\subsubsection{Text-Image Retrieval Module}

\label{sec:tir_short}
 \textbf{Relevant visual contexts $V$} (a subset of $P$) with respect to key phrases $Q$ can be identified with a Text-Image Retrieval Module $M_{\text{text-img}}$. This module ensures that the image regions most relevant to the textual contents are selected, enhancing the model's ability to focus on meaningful image areas for the given query.

 After the input image and text query are processed by $F$ and $G$, $n$ key phrases ($Q$) and $m$ image patches ($P$) will be encoded to text and image embeddings (d-dimension vectors) through a text encoder $f_\text{text}$ and an image encoder $f_{\text{img}}$, respectively. 
The similarity matrix between $n$ text embeddings and $m$ image embeddings is denoted as $S_\text{fast}$ ($S_\text{fast} \in \mathbb{R}^{n \times m}$) by calculate the cosine similarities between the embeddings of key phrases $Q$ and the embeddings of image patches $P$. Then, the cosine similarities will be normalized to $[0, 1]$ row-wise with the softmax function and a scale factor $\gamma$ \cite{clip}. 
\begin{align}
    M_{\text{text}} = f_{\text{text}}(Q), \text{where}\ M_{\text{text}} \in \mathbb{R}^{n \times d}
\end{align}
\begin{align}
    M_\text{img} = f_{\text{text}}(Q), \text{where}\ M_\text{img} \in \mathbb{R}^{n \times d}
\end{align}
\begin{align}
\label{eq:Sfast}
    S_\text{fast} = \text{Softmax}(\gamma \cdot M_{\text{text}}\ @ \ M_\text{img}^{\text{T}})
\end{align}
where $@$ represents matrix multiplication and \text{T} represent the matrix transpose. 

Each entry of the $S_\text{fast}$ matrix is in fact a similarity measurement that indicates how close an image patch and a text keyphrase are in the embedding space. Visual cues $V$ will be selected based on similarity matrix $S_{\text{fast}}$ using a selection function $H_{\text{fast}}$ with threshold $\epsilon$. 
\begin{align}
     V = H_{\text{fast}}(P, S_\text{fast}, \epsilon) = \{v_{i}\}_{i=1}^{k} 
\end{align}
where $k$ means there are $k$ patches satisfied the condition (e.g. patches with confidence greater than threshold $\epsilon$) and selected as visual cue. The \textbf{confidence $c_i$ for each visual cue $v_{i}$} is determined by looking up the $S_\text{fast}$ matrix finding the corresponding similarity.

If $k>0$, the selected $k$ visual cues will send to the MLLM directly for answer generation, which we call "\textbf{fast path}".

\subsubsection{Text-Text Retrieval Module and Vector Database}
\label{sec:t2trvd}
If $k=0$, a more complicate "\textbf{slow path}" will be proceed. First, we introduce the Text-Image Vector Database $D$, which stores million-scale labeled RSI with the key-value pairs as follows: the key is the \textbf{text of the class name or the image caption}, and the value is the \textbf{image embeddings} obtained using the image encoder $f_{\text{image}}$ with the set of images associated with that class or caption.

Given a set of query key phrases $Q = \{t_{i}\}_{i=1}^n $, $D$ retrieves the corresponding labels or captions $L=\{l_{p}\}_{p=1}^k$ whose text embeddings is close to the query text embedding $M_{\text{text}}$ (i.e. distance between two text embeddings is below certain threshold $\delta$). Formally, the retrieval process can be expressed as:

\begin{align}
\label{eq:database}
   L = \{l_{p}\}_{p=1}^k = D(Q, \delta)
\end{align}


where $l_{p}$ represents a label or caption in the database related to at least one key phrase from $Q$, and $\delta$ is a distance threshold. For each label or caption $l_{p}$, there is an associated collection of $s$ images $E_p = \{e_{i}^{p}\}_{i=1}^s$ from the vector database. We then use a proxy selection function $g$ to select the proxy image embedding among the $s$ image embeddings. The proxy image embedding can be expressed as following:

\begin{align}
\label{eq:proxy}
   \Tilde{E_p} = g(E_p) = g(\{e_{i}^{p}\}_{i=1}^s)
\end{align}

In summary, a set of key phrases $Q$ will form a set of relevant visual concepts $E$, where 
\begin{align}
E = \{\Tilde{E_p}\}_{p=1}^k
\end{align}

\label{sec:imageragmotivation}
\textbf{The motivation behind this step is to address the potential limitations of the Text-Image Retrieval Module} $M_{\text{text-img}}$, \textbf{which may not fully understand the visual concept of certain phrases in the domain of remote sensing} since $M_{\text{text-img}}$ could utilize a general VLM rather than RS-specialized one. For instance, the appearance of an aircraft can vary significantly depending on the perspective, such as a main view versus an satellite view.  \textbf{Fast path failure indicates that no visual concept has been found for the key phrases with high confidence.} However, it is unlikely that a text query would target a "void" concept since they are proposed by user. A plausible explanation is that Text-Image Retrieval Module $M_{\text{text-img}}$ utilizes VLM that is trained on general image-text pairs, making VLM is difficult to associate visual concepts in the RS domain with the textual descriptions. To address this, the \textbf{slow path leverages the text embeddings of the phrases and labels (or captions) as anchors and retrieves image embeddings from the RS domain within the database for phrase concepts}. These retrieved image embeddings serve as visual evidence and can be used for image-to-image search later, thereby enhancing the model's understanding of domain-specific visual concepts.

\subsubsection{Image-Image Retrieval Module} When visual evidence $E$ are obtained, we can calculate the similarity matrix of image patches $P$ and visual evidence $E$ just like equation \ref{eq:Sfast}.

\begin{align}
    S_\text{slow} = \text{Softmax}(\gamma \cdot E \ @ \ {M_\text{img}}^\text{T})
\end{align}

Visual context $V$ for \textbf{slow path} will be selected based on similarity matrix $S_{\text{slow}}$ using the selection function $H_{\text{slow}}$.
\begin{align}
     V = H_{\text{slow}}(P, S_\text{slow}) = \{v_{i}\}_{i=1}^{k} 
\end{align}

\subsection{Generation Stage}
Once visual cues $V$ are selected from $P$, a set of image patches from image $I$, MLLM can use such visual contexts for response generation. Unlike Ordinary RAG, which can directly organize retrieved text content with a prompt and send to a LLM to generate the response, ImageRAG must handle visual context. This means ImageRAG needs to select a MLLM that can utilize the visual contexts as visual cues. 

We designed a training set and trained a MLLM for this propose to make it be able to accept additional visual cues to focus on. The final response $R_i$ for given image $I_i$ and text query $T_i$ with visual cues $V_i$ and specifically designed prompt will be calculated following equation \ref{eq: MMLM_cues}. 

\section{Implementation Detail for ImageRAG}
\label{sec:implementationdetail}
In this section, we will outline the settings such as choices of models, parameters, training recipes, etc., along with the reason behind these choices.

\subsection{Patch Division Algorithm}
\label{sec:pd}


We select three patch division algorithms to patchify the images. The first is the Vision Transformer \cite{dosovitskiy2021imageworth16x16words} style ("\textbf{ViT}"), which partitions images into fixed-size, non-overlapping patches of size m $\times$ m. The second is a much denser division approach called "\textbf{Complete Cover}" patch from DetailCLIP \cite{zhang2023injectingimagedetailsclips}, which aims to partition images so that patches can cover objects of any scale, resulting in numerous non-overlapping multiscale image patches. The third is a compromise between the former two: a cascade approach ("\textbf{Cascade Grid}") that divides UHR RSIs into 1 $\times$ 1, $\cdots$, n $\times$ n grids. This results in overlapping patches at different scales compared to the ViT style, but not as densely as the DetailCLIP style.

For ViT-style patch division, we set m to 448 $\times$ 448, which matches the input size of InternVL2.5. For DetailCLIP-style division, we set the scale parameter to $\frac{a}{c} = 20$. This algorithm divides each image into overlapping, multiscaled image patches. The smallest patch size is approximately 200 $\times$ 200 pixels, and about 600 patches are generated per image. For the "Cascade Grid" approach, we set n to 10, resulting in 385 patches per UHR RSI.

\subsection{Question Analyzing Module}
\label{sec:qam}
This module is crucial for the ImageRAG framework, as it extracts the text of the key elements from the given question, which we plan to use consistently in later modules.  

The Question Analyzing Module G is implemented using \textbf{Qwen2.5-32B} \cite{qwen2.5}, with the $temperature$ set to 1.0, $top\_p$ set to 0.99, and $max\_tokens$ set to 512. The input prompt can be found in Appendix \ref{sec:appendix_prompt_qam}. We utilize SGLang \cite{zheng2024sglangefficientexecutionstructured} to host QWen2.5-32B for API serving. If a question fails to be parsed by QWen2.5-32B more than 10 times, we switch the Question Analyzing Module to \textbf{KeyBERT} \cite{grootendorst2020keybert}, with an n-gram range from 2 to 4 and a maximum of 3 phrases. We found that this implementation can already parse the question ideally, therefore, no extra technique was applied.

We asked the Question Analyzing Module to extract important keywords or phrases that include adjectives, while avoiding standalone adjectives or phrases related to position and orientation since we found they will confuse the MLLM by hacking the content of choices. Additionally, we aimed to exclude overly vague words such as "image", "picture", "photo", etc.

\subsection{Text-Image Retrieval Module}
\label{sec:tir}
\subsubsection{Selection of Image and Text Encoder}

\label{sec:encoder}
We select image and text encoders from four CLIP-based models: CLIP \cite{clip}, RemoteCLIP \cite{remoteclip}, GeoRSCLIP \cite{rs5m}, and MCIPCLIP \cite{schall2024optimizingclipmodelsimage}. The rationale is as follows:
\begin{itemize}
    \item \textbf{CLIP}: A classic, generalized model suitable for baseline tasks.
    \item \textbf{RemoteCLIP}: A satellite imagery specialized CLIP variant, making it appropriate for Remote Sensing benchmarks.
    \item \textbf{GeoRSCLIP}: It is trained not only on satellite view data but also on a large amount of aerial view data, providing greater diversity in remote sensing tasks.
    \item \textbf{MCIPCLIP}: While CLIP effective at text-image retrieval, it struggles to differentiate between visually distinct images with similar captions, leading to suboptimal performance in image-based similarity searches. To address this, MCIPCLIP uses an ArcMargin loss \cite{8953658} and an Multi-Caption-ArcMargin loss \cite{schall2024optimizingclipmodelsimage}, enhancing its image-image retrieval capabilities through metric learning.
\end{itemize}

\subsubsection{The Similarity Matrix}
\label{sec:fastpathsim}
 As mentioned in section \ref{sec:tir_short}, the similarity matrix $S_{\text{fast}}$ between $n$ key phrases and $m$ image patches is calculated using text and image embeddings encoded by corresponding encoder. Key phrases are prompted with the CLIP template \footnote{https://github.com/openai/CLIP/blob/main/notebooks/ \\Prompt\_Engineering\_for\_ImageNet.ipynb} and the average text embedding is taken. A sentence that includes all key phrases along with the entire image as an image patch are added separately.

\subsubsection{Patch Selection Function}
Given $S_{\text{fast}}$ ($S_{\text{fast}} \in \mathbb{R}^{n \times m}$), $H_\text{fast}$ first selects the 2 most frequently appearing image patches across all key phrases. Then, for each key phrase, the image patch with the highest similarity is selected. After removing duplicate image patches, the \textbf{top-5 image patches with high similarity are retained}. \textbf{The similarity score, ranging from 0 to 1, for each image patch serves as the confidence measure for that patch.} If no image patch is selected with a confidence score above $\epsilon$ (set to 0.5), the slow path will be triggered. Otherwise the image patches with confidence above $\epsilon$ will be returned.

\subsection{Text-Text Retrieval Module and Vector Database}
\label{sec:ttrm}

Two Text-Image Vector Databases are set up: one that contains class-wise labeled images related to remote sensing, referred to as the Labeled Remote Sensing Database (LRSD), and another that includes RS-related images with captions, named Captioned Remote Sensing Database (CRSD).  Unlike the simplified definitions in Section \ref{sec:t2trvd} and equation \ref{eq:database}, we now expand the definitions definition in practice from $D$ to $D_1$ and $D_2$ and adjust the distance threshold from $\delta$ to the respective $\delta_{1}$ and $\delta_{2}$ for each vector database.

\subsubsection{LRSD}
We collect images of different instances by cropping objects from commonly used remote sensing datasets, including those for classification, object detection, and segmentation. DOTAv2.0\cite{dota2}, FAIR1M \cite{sun2021fair1mbenchmarkdatasetfinegrained}, iSAID\cite{isaid}, SODA-A \cite{sodaa}, LoveDA \cite{loveda}, MillionAID \cite{millionaid}, and FMoW \cite{fmow} are selected. For classification datasets, we use the entire image. For object detection datasets, we crop the bounding box with a zoom-out ratio of 1.3 (to include some background of the object). For segmentation datasets, we convert the segmentation mask to a detection box, and cropping follows the same method as for detection datasets. We only crop objects from the training set of each dataset, and a deduplication process with d-hash \footnote{https://github.com/benhoyt/dhash} is applied. 230,958 duplicate images are removed. Table \ref{table:LRSD} shows the details of LRSD. There are images from 142 classes in the LRSD. List of class names can be found in Appendix \ref{sec:appendix_LRSD}.

\begin{table}[ht]
\centering
\caption{LRSD Per Dataset Statistics. Average Object Size is in pixels.}
\label{table:LRSD}
\begin{tabularx}{\linewidth}{>{\centering\arraybackslash}X c c c}
\toprule
\textbf{Dataset} & \textbf{Task Type} & \textbf{Count} & \textbf{Average Object Size} \\
\midrule
FMoW             & Detection   &  363,572 & 357.25 $\times$ 282.84 \\
DOTAv2.0         & Detection   & 261,277 & 33.56 $\times$ 33.54 \\
FAIR1M        & Detection   & 213,107 & 26.70 $\times$ 24.30 \\
iSAID        & Segmentation   & 321,235 & 23.79 $\times$ 24.34 \\
SODA-A           & Segmentation & 337,336 & 22.17 $\times$ 21.96 \\
LoveDA           & Segmentation & 92,254  & 161.88 $\times$ 148.98 \\
MillionAID       & Classification & 7,128 & 543.17 $\times$ 543.17 \\
Total           & -   & 1,595,909 & - \\
\bottomrule
\end{tabularx}
\end{table}

\subsubsection{CRSD} 
We use PUB11 subset from RS5M dataset as the CRSD due to its diversity. There are \textbf{3,007,809} RS related image-text pairs collected from 11 public large-scale English image-text paired datasets, including LAION2B-en \cite{laion5b}, LAION400M \cite{laion400m}, LAIONCOCO, COYO700M \cite{coyo700m}, CC3M \cite{cc3m}, CC12M \cite{cc12m}, YFCC15M \cite{yfcc100m}, WIT \cite{wit}, Redcaps \cite{redcaps}, SBU \cite{sbu}, and Visual Genome \cite{vg}.

\subsubsection{Retrieval Process}
 
Chroma \footnote{https://github.com/chroma-core/chroma} is chosen as our vector database. Label and caption information from LRSD and CRSD are encoded using Sentence-Bert \cite{reimers-2019-sentence-bert} ("all-MiniLM-L6-v2" model) and stored in Chroma. The $L_2$ distance is employed as the metric for measuring the distance between two text embeddings in the vector database.

When key phrases are routed to the slow path, their text embeddings (generated by Sentence-Bert) are computed. These embeddings are then used to search the LRSD for labels with a close match (i.e., an $L_2$ distance below $\delta_1$, with $\delta$ set to 0.3). If such labels exist, they and their corresponding images are returned and the proxy image embedding will be calculated. If no matches are found in LRSD, the text embeddings are used to search the more diverse but noisier CRSD, returning candidates with an $L_2$ distance below $\delta_2$ (set to 0.5).

\subsubsection{Proxy Selection Function}
As we mentioned in section \ref{sec:t2trvd} and equation \ref{eq:proxy}, we need to calculate proxy image embeddings that represent the visual concepts for each class in the LRSD database for image-image retrieval. Three approaches are implemented, which are "Prototype", "Clustering", and "Reranking".

\begin{itemize}
    \item \textbf{Prototype}: Following the approach of ProtoNet \cite{snell2017prototypicalnetworksfewshotlearning}, all image embeddings are normalized first and the mean image embedding is computed from all images that share the same class label. This mean image embedding is used as the proxy image embedding of the class.
    \item \textbf{Clustering}: We first cluster the image embeddings within the same class using Density-Based Spatial Clustering (DBSCAN)\cite{dbscan} (scikit-learn implementation \footnote{https://scikit-learn.org/stable/modules/generated/sklearn.cluster.DBSCAN.html}). We set epsilon to 0.3 and \text{min\_samples} to 5. Then, we select the largest cluster and calculate the mean image embeddings of all images belonging to it. This mean image embedding serves as the proxy for visual concept representation in its respective class.
    
    \item \textbf{Reranking}: Reranking is a key technique for enhancing the quality and relevance of search results in RAG. The process involves taking the initial set of data retrieved based on a user's query and reordering it using advanced models that can better understand context and semantic meaning. To implement reranking, we first use the class label in the vector database to locate the original key phrase. We then take the corresponding text feature from the fast path. Next, we calculate the similarity matrix between the key phrase text feature and the image features within the label class. We select the top 3 image features and compute their mean to use as the proxy image feature. 

\end{itemize}

\subsection{Image-Image Retrieval Module}

Similar to the Text-Image Retrieval Module in the fast path from section \ref{sec:fastpathsim}, $S_\text{slow}$ is calculated in the Image-Image Retrieval Module. However, unlike $S_\text{fast}$, which measures the similarity between text key phrases and image patches, $S_\text{slow}$ calculates the similarity between visual evidence (proxy image features) and image patches. The ranking process and confidence calculation are consistent with those in the fast path. $H_\text{slow}$ selects 3 image patches with the highest confidence.

\subsection{Multimodal Large Language Model with Visual Cues}
\label{sec:mllmvisualcue}

Despite their promising capabilities, recent advanced MLLMs often struggle to link the positional relationships between the global image and sub-images, where sub-images are an array of sub-patches from the identical global image~\cite{shen2024zoomeye}. This means that when these visual contents are input collectively, the current MLLMs cannot accurately localize the position of the sub-images within the global image. In other words, MLLMs cannot directly infer the answer based on visual cues. Moreover, general instruction-tuned MLLMs tend to exhibit suboptimal performance in specialized professional domains, such as Remote Sensing \cite{vrsbench} \cite{skysensegpt} and Medical Imaging \cite{bai2024m3dadvancing3dmedical} \cite{AlSaad2024-pc}. Consequently, it is unwise to integrate an open-sourced MLLM into our framework without modification.


To address these two issues, we curated a global-local \textbf{visual-cue-aware dataset} specialized for the remote sensing domain to fine-tune general MLLMs, which we call \textbf{Zoom4K}. Specifically, we leveraged the bounding-box annotations of an $instance$ in the global image from existing off-the-shelf RS datasets. A 512 $\times$ 512 pixels sub-patch containing this $instance$ and $k$ other arbitrary sub-patches were first cropped from the global image (use as distractor for robustness consideration). Denoting the bounding boxes of these sub-patches as $\{b_{gt}, b_1, \dots, b_k\}$, $I$ as the global image, and $I_{bi}$ as the sub-images cropped from $I$ corresponding to $b_i$, we then combined $(I,\ \mathcal{R}(\{(b_i,I_i)\}_{ i\in\{gt,1,\dots,k\} }),\ Q_{pos})$ as the MLLM input. Here, $Q_{pos}$ represents a question querying the position of the $instance$ (e.g., "\texttt{Where is the $\{instance\}$ in the picture?}"), and $\mathcal{R}(\cdot)$ is a function used to randomly shuffle the elements within the array. Finally, the position of the $instance$ was viewed as the target output of the MLLM.

\begin{figure}[t]
    \centering
    \includegraphics[width=0.49\textwidth]{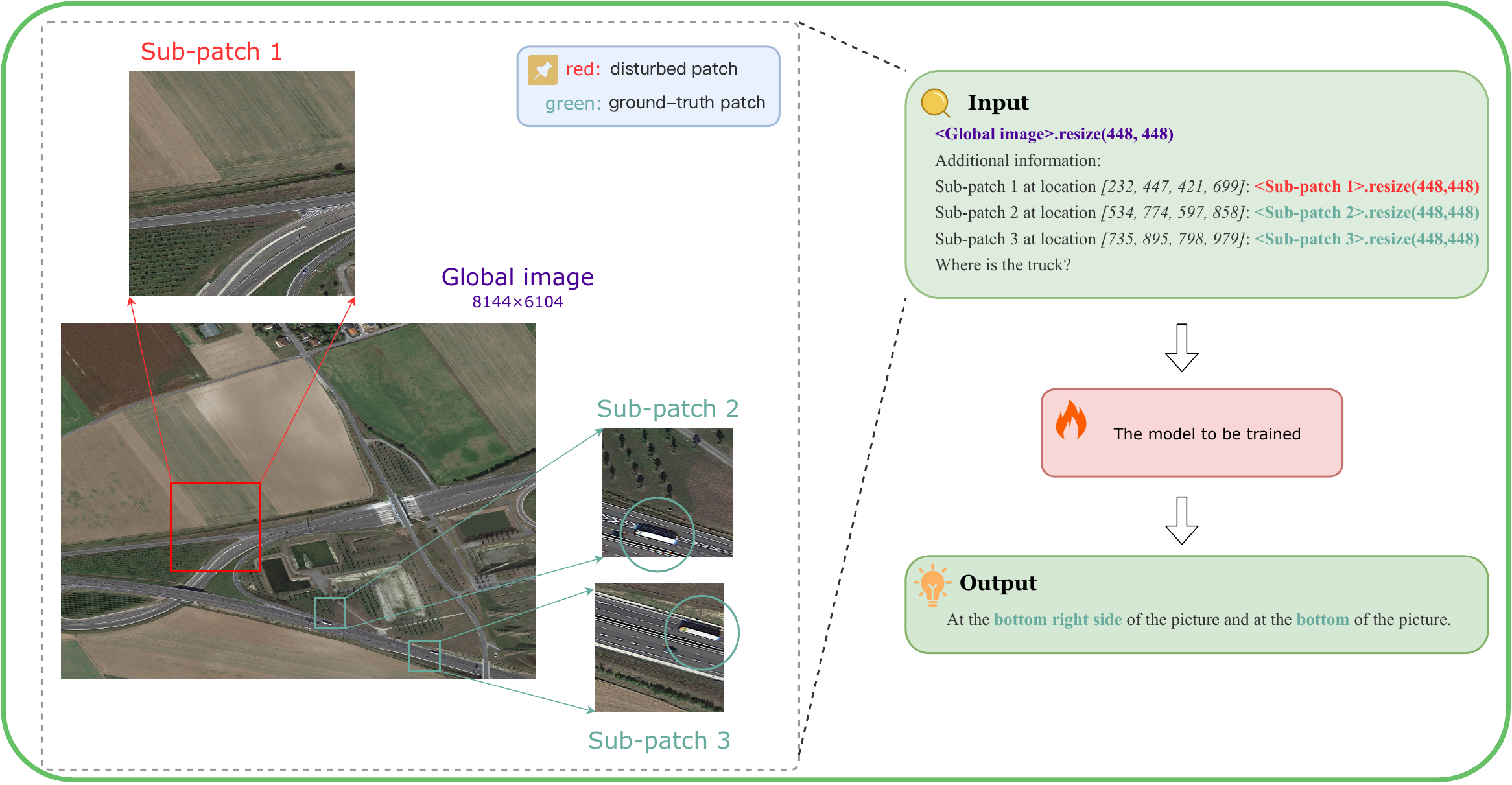}
    \caption{A demonstration of \textbf{Zoom4K} and how to fine-tune Multimodal Large Language Model with visual cues}
    \label{fig:zoom4k}
\end{figure}

We categorize the position output as follows: \textit{center, top, bottom, left, right, top-left, top-right, bottom-left, bottom-right, center-left, center-right}. These categories are determined based on the center coordinates of the bounding box annotation of the $instance$, as illustrated in Figure \ref{fig:zoom4k}. The position data, totaling approximately 4,000 entries, is sourced from the training sets of DOTA2 \cite{dota2}, SODA \cite{sodaa}, and FIT \cite{skysensegpt} (linked to original images in STAR \cite{li2024starfirsteverdatasetlargescale}). All data are UHR RSI. \textbf{Zoom1K} is a subset of Zoom4K that only uses STAR as its data source.


To further enhance the model's capabilities, we curated an additional 10,000 VQA dataset (\textbf{VQA10K}) for model training, aiming to strengthen the model's global-local understanding while maintaining its VQA capability. Specifically, we used STAR \cite{li2024starfirsteverdatasetlargescale} and FIT \cite{skysensegpt} to construct the VQA data. These datasets divide an UHR RSI into a series of small images, each with $512 \times 512$ pixels, and annotate each one with several VQA samples based on its individual visual content. Denoting the original UHR image as $I$, the bounding box of the tiny image located in $I$ as $b$, and the visual content of the tiny image as $I_b$, we combine $(I,\ (b,I_b),\ Q)$ as the MLLM input, where $Q$ represents the annotated question in STAR, and the corresponding annotated answer $A$ is treated as the target output. We then used the next-token-prediction loss \cite{vaswani2023attentionneed} to train the MLLM on these two constructed position and VQA datasets. The fine-tuned model obtained through this process is denoted with the suffix \textbf{-Infer} in our paper.

Through this training process, we achieve two key objectives: (1) We endow the model with the ability to perceive the position of sub-images within the global image, creating an inferring model that is visual-cue-aware; (2) We inject remote sensing domain knowledge into the model.

We used Low-Rank Adaptation (LoRA) \cite{hu2021loralowrankadaptationlarge} with an alpha value of 32 to fine-tune the InternVL2.5-8B model on a mixed dataset of Zoom4K and VQA10K. This process resulted in the InternVL2.5-8B-Infer, a model for inferring VQA task.

\section{Experiment}
\subsection{Experiment Setup}
\label{sec:mainexp}

In this section, we present the experimental results for the tasks introduced in Section \ref{sec:task}. We select InternVL2.5-8B \cite{internvl25} as our base model, and MME-RealWorld-lite-RS (introduced in Section \ref{sec:MME-RealWorld-Lite-RS}) as the primary benchmark. Other settings are detailed in Section \ref{sec:implementationdetail}. InternVL2.5-8B-Infer is the \textbf{MLLM with visual cues} introduced in section \ref{sec:mllmvisualcue}. The prompt template for inferring model can be found in Appendix \ref{appendix:pt_inferringmodel}.

Unless specified, we set $\epsilon$ to 0.5, $\delta_1$ to 0.3, and $\delta_2$ to 0.5. The Patch Division Algorithm is chosen as "Cascade Grid", the Image and Text Encoder are selected from "CLIP", and the Proxy Selection Function adopts "Clustering". The random seeds across all experiments are set to 2024.

\subsection{Result of Regular VQA task}
\label{section:main_result}

\begin{table*}[htbp]
\caption{Experimental results of \textbf{Regular VQA Task} on the MME-RealWorld-lite-RS. "Position", "Color", and "Count" each indicate a specific sub-task domain mentioned in section \ref{sec:MME-RealWorld-Lite-RS}. The model with the "-Infer" suffix is an inferring model. MGM-7B is fine-tuned with VRSBench training data. "5-epoch" means the model has been fine-tuned for five epochs. \textcolor{black}{"High-Res Model" means the high-resolution vision encoder is applied, and "Native-Res" is the abbreviation for "Native-Resolution". "MoE" stands for Mixture-of-Experts architecture.}   "ROI Box (GT)" indicates the use of ground truth Region-of-Interest boxes as visual cues. Inferring models using ROI boxes can be seen as the upper bound of the ImageRAG framework.}
\label{tab:res_main_perception}
\centering
\begin{tabular}{ccccccccc}
\toprule 
\Gray
\multicolumn{1}{c}{\textbf{Model}} & \multicolumn{1}{c}{\textbf{LLM}} &\multicolumn{1}{c}{\textbf{Image Encoder}} & \multicolumn{1}{c}{\textbf{Image Resolution}}  & \multicolumn{1}{c}{\textbf{Note}} & \multicolumn{4}{c}{\textbf{Accuracy}}  \\ \Gray
\multicolumn{5}{c}{\textbf{Task Split}}  & \textbf{Position} & \textbf{Color} & \textbf{Count} & \textbf{Average} \\  
\Gray
\midrule
\multicolumn{9}{c}{\textbf{MME-RealWorld-Lite-RS}} \\
LLaVA1.5 \cite{liu2023improvedllava} & Vicuna1.5-7B & CLIP-ViT-L-14-336px & 336 $\times$ 336 & - & 24.00&26.00  &18.00 & 22.67\\
MGM \cite{li2024minigeminiminingpotentialmultimodality} & Vicuna1.5-7B & CLIP-ViT-L-14-336px & 336 $\times$ 336 & VRSBench\cite{vrsbench} + LoRA& 20.00 & 28.00 & 28.00 & 25.33 \\
VHM \cite{H2RSVLM} & Vicuna1.5-7B & CLIP-ViT-L-14-336px & 336 $\times$ 336 & - & 28.00 & 26.00 & 28.00 & 27.33\\
Geochat \cite{geochat} & Vicuna1.5-7B & CLIP-ViT-L-14-336px & 504 $\times$ 504 & - & 22.00 & 36.00 & 26.00 & 28.00 \\
SkysenseGPT \cite{skysensegpt} & Vicuna1.5-7B & CLIP-ViT-L-14-336px & 504 $\times$ 504 & - & 30.00 & 38.00 & 30.00 & 32.67 \\
\midrule
\textcolor{black}{LLaVA-HR} \cite{llavahr} & \textcolor{black}{Vicuna1.5-7B} & \textcolor{black}{CLIP-ViT-L-14-336px} & \textcolor{black}{1024 $\times$ 1024} & \textcolor{black}{High-Res Model} & \textcolor{black}{32.00} & \textcolor{black}{38.00} & \textcolor{black}{16.00} & \textcolor{black}{28.67} \\
\textcolor{black}{LLaVA-UHDv2} \cite{zhang2024llavauhdv2} & \textcolor{black}{Qwen2.0-7B} & \textcolor{black}{CLIP-ViT-L-14-336px} & \textcolor{black}{672 $\times$ 1008} & \textcolor{black}{High-Res Model} & \textcolor{black}{46.00} & \textcolor{black}{32.00} & \textcolor{black}{10.00} & \textcolor{black}{29.33} \\
\textcolor{black}{Kimi-VL-A3B} \cite{kimiteam2025kimivltechnicalreport} & \textcolor{black}{MoE-16B} & \textcolor{black}{MoonViT-SO-400M} & \textcolor{black}{Native-Res} & \textcolor{black}{High-Res Model} & \textcolor{black}{58.00} & \textcolor{black}{48.00} & \textcolor{black}{16.00} & \textcolor{black}{40.67} \\
\midrule
InternVL2.5-8B \cite{internvl25} & Qwen2.5-7B & InternViT-300M-V2.5 & 448 $\times$ 448 & Vanilla Model Baseline & 56.00 & 54.00 & 30.00 & 46.67 \\
InternVL2.5-8B-Infer & Qwen2.5-7B & InternViT-300M-V2.5 & 448 $\times$ 448 & Inferring Model Baseline & 54.00 & 54.00 & \textbf{32.00} & 46.67 \\
InternVL2.5-8B-Infer & Qwen2.5-7B & InternViT-300M-V2.5 & 448 $\times$ 448 & 5-epoch & 52.00 & 52.00 & 30.00 & 44.67  \\
\Lgray InternVL2.5-8B-Infer & Qwen2.5-7B & InternViT-300M-V2.5 & 448 $\times$ 448 & \textbf{ImageRAG} & \textbf{64.00} & \textbf{62.00} & 30.00 & \textbf{52.00} \\
\midrule
InternVL2.5-8B \cite{internvl25} & Qwen2.5-7B & InternViT-300M-V2.5 & 448 $\times$ 448 & ROI Box (GT) & 56.00 & 68.00 & 42.00 & 55.33 \\
InternVL2.5-8B-Infer & Qwen2.5-7B & InternViT-300M-V2.5 & 448 $\times$ 448 & ROI Box (GT) & 58.00	&78.00&	44.00 & 60.00 \\

\bottomrule
\end{tabular}%
\end{table*}
We compare the InternVL2.5-8B-Infer model with other well known RSMLLMs including Mini-Gemeni \cite{li2024minigeminiminingpotentialmultimodality} (fine-tuned in VRSBench \cite{vrsbench} using LoRA), VHM-7B \cite{H2RSVLM}, Geochat \cite{geochat}, SkysenseGPT \cite{skysensegpt}. 
\textcolor{black}{Competitive high-resolution MLLMs are compared as well, such as LLaVA-HR \cite{llavahr}, LLaVA-UHDv2 \cite{zhang2024llavauhdv2}, and Kimi-VL-A3B-Instruct \cite{kimiteam2025kimivltechnicalreport} (An 16B Mixture-of-Experts MLLM with 2.8B activate parameters released in April 2025. Notably, its vision encoder supports native resolution.)}
All compared models use their individual conversation template and the question template from MME-RealWorld dataset \cite{mmerealworld} (can be found in Appendix \ref{appendix:pt_mmerealworld}. Table \ref{tab:res_main_perception} shows the result of different MLLMs and InternVL2.5-8B-Infer model in MME-RealWorld-lite-RS dataset. Base LLM, image encoder, supported image resolution are listed.

\textcolor{black}{The InternVL2.5-8B model exhibits remarkable performance in the Regular VQA task on the MME-RealWorld-lite-RS dataset, significantly surpassing other MLLMs with an average accuracy increase of 6\% to 24\%. As detailed in Table \ref{section:main_result}, it is evident that both fine-tuning MLLMs with remote sensing SFT data (data-centric approach) and enhancing the vision encoder's supported image resolution of MLLMs (model-centric approach) are equally effective for the Regular VQA task involving high-resolution remote sensing images. When compared to LLaVA1.5, an MLLM lacking support for high-resolution image input and fine-tuning with remote sensing data, the data-centric approach, such as that employed by Geochat and SkysenseGPT, can elevate results from 22.67\% to 32.67\%. On the other hand, the model-centric approach, as seen in LLaVA-HR and LLaVA-UHDv2, can enhance results from 22.67\% to 29.33\%. Furthermore, a more powerful LLM backbone is also vital. For instance, Kimi-VL-A3B, equipped with a 16B MoE language model, and InternVL2.5-8B, with a Qwen2.5-7B language model, both demonstrate an average accuracy of over 40\%.}

The InternVL2.5-8B-Infer model (inferring model baseline), with no ground truth ROI boxes provided, maintains performance comparable to the vanilla InternVL2.5-8B model baseline (46.67\% v.s. 46.67\%). This suggests that our fine-tuning process to develop an inferring model capable of reasoning with visual cues does not compromise the integrity or effectiveness of the original model. However, when the model is fine-tuned for more epochs (e.g. 5 epochs), there is a noticeable decline in performance ({\bf \textcolor{emerald!80}{-2\%}}). This contrasts with results presented in Table \ref{tab:inferringvqa}, where even with 5 epochs of fine-tuning, the model shows improved performance ({\bf \textcolor{coralred!80}{+2\%}}) when provided with ground truth ROI boxes. This indicates that additional training steps may cause the model to become more reliant on the provided visual cues, potentially reducing its ability to generalize or reason effectively without accurate ROI information.


The InternVL2.5-8B-Infer model, when provided with ground truth ROI boxes, establishes a \textbf{performance upper bound} for this task (60\%). As shown in Table \ref{tab:res_main_perception}, the application of the ImageRAG method leads to a significant performance boost, raising the accuracy from a baseline (both vanilla model and inferring model) of 46.67\% to 52\% ({\bf \textcolor{coralred!80}{+5.33\%}}). This substantial improvement highlights the effectiveness of the ImageRAG technique in enhancing the model's ability to reason with retrieved visual cues. However, despite this notable enhancement, there remains a gap between the current performance and the upper bound set by the ground truth ROI box. This indicates that while ImageRAG is effective, there is still room for improvement and optimization to reach the full potential of the model. 

\subsection{Result of Inferring VQA Task}
\label{sec:infer_vqa}
\begin{table*}[htbp]
\caption{Experimental results of \textbf{Inferring VQA Task} on the MME-RealWorld-lite-RS. "Position", "Color", and "Count" each indicate a specific sub-task domain mentioned in section \ref{sec:MME-RealWorld-Lite-RS}. The model with the "-Infer" suffix is an inferring model. "5-epoch" means the inferring model has been fine-tuned for five epochs. "ROI Box (GT)" indicates the use of ground truth Region-of-Interest boxes as visual cues. "Random Box" means a random box from UHR image is given for the model. Zoom1K, Zoom4K, and VQA10k are the training set mentioned in section \ref{sec:mllmvisualcue}, and "CoT" represents the Chain-of-Thought technique.}
\label{tab:inferringvqa}
\centering
\begin{tabular}{cccccccc}
\toprule \Gray
\multicolumn{1}{c}{\textbf{Model}} & \multicolumn{1}{c}{\textbf{Technique}} & \multicolumn{1}{c}{\textbf{Fine-tune Data}} & \multicolumn{1}{c}{\textbf{Image Resolution}} & \multicolumn{4}{c}{\textbf{Accuracy}}  \\ \Gray
\multicolumn{4}{c}{\textbf{Task Split}}  & \textbf{Position} & \textbf{Color} & \textbf{Count} & \textbf{Average} \\  
InternVL2.5-8B \cite{internvl25} & Vanilla Model Baseline & - & 448 $\times$ 448 & 56.00 & 54.00 & 30.00 & 46.67 \\
InternVL2.5-8B-Infer & Inferring Model Baseline & Zoom4K + VQA10K & 448 $\times$ 448 & 54.00 & 54.00 & 32.00 & 46.67 \\
InternVL2.5-8B \cite{internvl25} & ROI Box (GT)& - & 448 $\times$ 448 & 56.00 & 68.00 & 42.00 & 55.33 \\
InternVL2.5-8B-Infer & ROI Box (GT) & Zoom1K & 448 $\times$ 448 &  52.00 &	76.00	&40.00& 56.00 \\
InternVL2.5-8B-Infer & ROI Box (GT) & Zoom4K & 448 $\times$ 448 & 54.00	& 72.00 & 46.00 & 57.33 \\
InternVL2.5-8B-Infer & ROI Box (GT) & Zoom4K + VQA10K & 448 $\times$ 448 & 58.00	&78.00&	44.00 & 60.00 \\
InternVL2.5-8B-Infer & Random Box & Zoom4K + VQA10K & 448 $\times$ 448 & 46.00 & 42.00 & 18.00 & 35.33 \\
InternVL2.5-8B-Infer & ROI Box (GT) + 5-epoch & Zoom4K + VQA10K & 448 $\times$ 448 & \textbf{68.00}	&76.00&	42.00 & 62.00 \\
InternVL2.5-8B-Infer & ROI Box (GT) + CoT & Zoom4K + VQA10K & 448 $\times$ 448 & 62.00	&\textbf{82.00} &	\textbf{50.00} & \textbf{64.67} \\		
\bottomrule
\end{tabular}%
\end{table*}

In Table \ref{tab:inferringvqa}, we provides a detailed comparison of various techniques for the inferring VQA task. Compared with vanilla model baseline, introducing ROI box (can be considered as an image patch contains question related visual cues) with text prompt in zero-shot manner already significantly boosts Color accuracy from 54\% to 68\% , while maintaining Position accuracy (56\%) and improving Counting accuracy to 42\%, resulting in an overall performance increase of {\bf \textcolor{coralred!80}{8.66\%}}. 
This indicates that using correct visual cues in a training-free model can already enhance its ability to understand color, position and counting questions, due to better localization of relevant visual features. 


Further fine-tuning with Zoom1K enhances Color accuracy (from 68\% to 76\%), but it negatively impacts the Position and Count tasks. This decline is reversed when the training data is increased from 1K to 4K. Additionally, incorporating VQA-like data (QA pairs with UHR RSI) significantly boosts all subtasks, achieving an average accuracy of 60\% ({\bf \textcolor{coralred!80}{+13.33\%}}). Adding the zero-shot Chain-of-Thought (CoT) \cite{wei2023chainofthoughtpromptingelicitsreasoning} reasoning technique yields the highest overall performance, with Position accuracy of 62\%, Color accuracy of 82\%, Count accuracy of 50\%, and an average of 64.67\% ({\bf \textcolor{coralred!80}{+18\%}}).

These results highlight the effectiveness of visual-cue-aware inferring model in improving VQA performance. The best-performing model demonstrates significant improvements across all subtasks, indicating that these techniques effectively address the challenges of the inferring VQA task. 

\subsection{Result of Visual Cue Retrieval Task}
In this section, we aim to verify if ImageRAG can retrieve useful visual cues by assessing the overlap between these cues and the ground truth ROI box. Unlike our approach in the main experiment (section \ref{section:main_result}), where we filtered visual cues based on confidence and $\epsilon$, potentially resulting in no visual cue output if the confidences were too low, we now force the model to output visual cues even if they do not meet the $\epsilon$ threshold.
Mean Recall is calculated following equation \ref{eq: mr}.  

\begin{table}[htbp]
\caption{Experimental results of \textbf{Visual Cue Retrieval Task} on the MME-RealWorld-lite-RS. "Mode" indicates ImageRAG use fast path only or mix of fast path and slow path mentioned in section \ref{sec:retrievalstage}. "Vector Database" represents the usage of LRSD and CRSD mentioned in section \ref{sec:ttrm}, and "Mean Recall" indicate the average of all Recall@3 from equation \ref{eq: mr}. "T" is the IoU threshold from equation \ref{eq: indicator}.}
\label{table:retrieve_visualcue}
\centering
\begin{tabular}{cccc}
\toprule \Gray
 \multicolumn{1}{c}{\textbf{Mode}} & \multicolumn{1}{c}{\textbf{Vector Database}} & \multicolumn{1}{c}{\textbf{MR (T=0.1)}} & \multicolumn{1}{c}{\textbf{MR (T=0.3)}} \\ 
Fast Path & - & 16.89 & 4.44\\
Fast + Slow Path & LRSD & 17.11 & 5.33 \\
Fast + Slow Path & CRSD & 17.11 & 5.11 \\
Fast + Slow Path & LRSD + CRSD & \textbf{17.33} & \textbf{5.33} \\
\bottomrule
\end{tabular}%
\end{table}


From Table \ref{table:retrieve_visualcue}, we can see that introducing the slow path mechanism boosts the mean recall for both T=0.1 and T=0.3, with even higher mean recall at T=0.1 when both LRSD and CRSD are used. However, the result is not significant because we force the model to output visual cues even if they do not meet the $\epsilon$ threshold, which hurt the result under this metric. However, if we did not use this metric, the comparison of the mean recall will not be fair. The reason is, in practice, we find that both the vanilla and inferring models are sensitive to the accuracy of the retrieved visual cues. Inaccurate visual cues can sometimes lead to worse ImageRAG performance than the baseline. In Table \ref{tab:inferringvqa}, we demonstrate this phenomenon: a noticeable ({\bf \textcolor{emerald!80}{-26.67\%}}) decline is happened when we replace the ground truth ROI box with a random box. Therefore, during deployment, we adopt a conservative strategy: we do not provide the visual cue if the retrieved result is not believable (\textbf{providing incorrect information is worse than providing none}). That is, we filter out visual cues with lower confidence, which enhances the overall framework's accuracy. 

When slow path is disabled, fast path is used less frequently (10.67\% utilization). The model tends to answer questions without ImageRAG, in a zero-shot manner. As fast path is uncertain, the framework often outputs void visual cues. However, when slow path is applied, the model uses ImageRAG more frequently to answer questions (45.33\% utilization).

\subsection{Result Robustness on Larger Dataset}

\textcolor{black}{In Section \ref{section:main_result}, ImageRAG showed remarkable performance on the MME-RealWorld-Lite-RS dataset. However, this dataset is relatively small, containing only 150 samples in total, with 50 samples per subtask. To test how ImageRAG performs with larger datasets, we created a series of datasets with different volumes.}

\textcolor{black}{We add 360 more samples (120 per subtask) from MME-RealWorld-RS, following the labeling standards outlined in section \ref{sec:MME-RealWorld-Lite-RS}. These new samples do not overlap with those in MME-RealWorld-Lite-RS and were examined by human to make sure the question and answer are not ambiguous nor vague. We refer to the original MME-RealWorld-Lite-RS dataset as $\mathbb{D}_{150}$. We also present $\mathbb{D}_{300}$, $\mathbb{D}_{450}$, and $\mathbb{D}_{510}$, each with an equal number of samples for each subtask. Moreover, each smaller dataset is a subset of the larger ones. This means $\mathbb{D}_{150} \subseteq \mathbb{D}_{300} \subseteq \mathbb{D}_{450} \subseteq \mathbb{D}_{510}$.}

\begin{table}[htbp]
\caption{Performance of ImageRAG on larger datasets.}
\label{table:resultrobustness}
\centering
\begin{tabular}{ccccc}
\toprule \Gray
 \multicolumn{1}{c}{\textbf{Dataset}} & \multicolumn{4}{c}{\textbf{Accuracy}}\\ 
 \Gray
 \multicolumn{1}{c}{\textbf{}} & \multicolumn{1}{c}{\textbf{Position}} & \multicolumn{1}{c}{\textbf{Color}} & \multicolumn{1}{c}{\textbf{Count}} & \multicolumn{1}{c}{\textbf{Average}} \\ 
$\mathbb{D}_{150}$ &  \textbf{64.00} & 62.00 & 30.00 & \textbf{52.00} \\
$\mathbb{D}_{300}$ &  54.00 & \textbf{65.00} & \textbf{33.00} & 50.67 \\
$\mathbb{D}_{450}$ & 58.67 & 63.33 & 32.00 & 51.33 \\
$\mathbb{D}_{510}$ & 59.41 & 64.12 & 32.35 & 51.96\\
\bottomrule
\end{tabular}
\end{table}

\textcolor{black}{As shown in Table \ref{table:resultrobustness}, ImageRAG demonstrates consistent performance across datasets of varying sizes, indicating that it is robust on larger datasets.
}

\section{Ablation Study}
\label{sec:ablation}
In this ablation study, we examine the key factors that influence ImageRAG's performance, such as the Patch Division Algorithm, the Image and Text Encoder, and the Proxy Selection Function. We use an ensemble of results choose from different thresholds to study: $\epsilon \in \{0.3, 0.5, 0.7\}$, $\delta_1 \in \{0.1, 0.2, 0.3\}$, and $\delta_2 \in \{0.3, 0.5, 0.7\}$. The rationale behind this setting is we aim to limit the return of inaccurate visual cues, particularly for $\delta_1$, as it controls the output from querying LRSD database. Additionally, we explore how the performance of the inferring model changes when the ROI size is enlarged, which may cause the evidence to become more vague.

\subsection{Encoding Models}
\begin{figure}[htbp]
    \centering
    \includegraphics[width=0.49\textwidth]{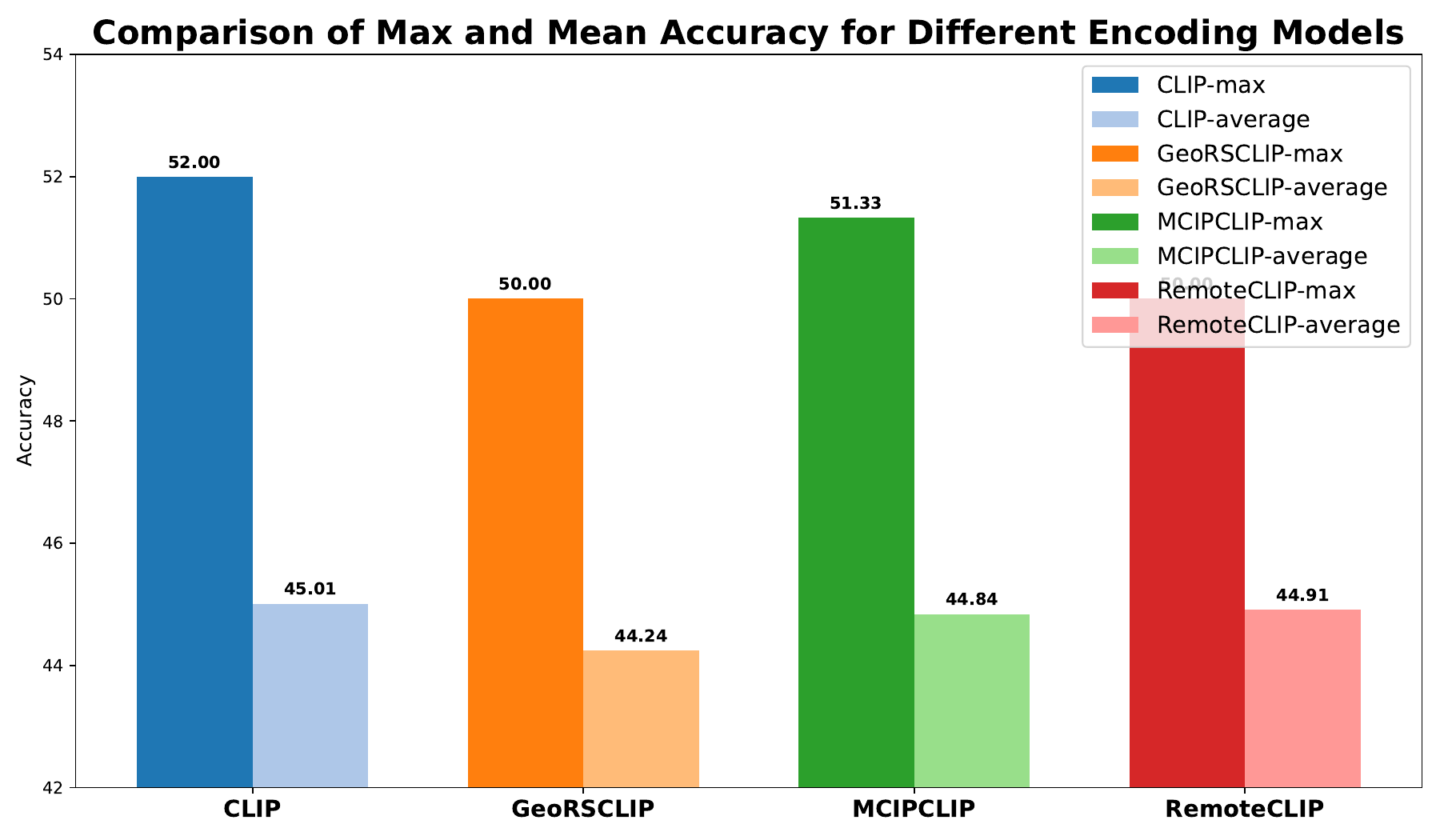}
    \caption{Comparison of max and mean accuracy for different encoding models}
    \label{fig:encoder}
\end{figure}

As mentioned in Section \ref{sec:encoder}, we chose CLIP, RemoteCLIP, GeoRSCLIP, and MCIPCLIP for comparison. Results are presented in Figure \ref{fig:encoder}. Notably, the CLIP model achieves the highest average and maximum accuracy, outperforming two models fine-tuned on remote sensing data (GeoRSCLIP and RemoteCLIP). Despite being designed for image-image retrieval task (crucial in the slow Path), MCIPCLIP does not show the expected performance improvement in average performance.

\subsection{Patch Division Algorithm}
\begin{table}[htbp]
\caption{Regular VQA Task result with different patch division algorithms.}
\label{table:patchdiv}
\centering
\begin{tabular}{ccccc}
\toprule \Gray
 \multicolumn{1}{c}{\textbf{Patch Division Algorithm}} & \multicolumn{4}{c}{\textbf{Accuracy}} \\ \Gray
  \multicolumn{1}{c}{Task Split} & \multicolumn{1}{c}{\textbf{Position}} & \multicolumn{1}{c}{\textbf{Color}} & \multicolumn{1}{c}{\textbf{Count}} & \multicolumn{1}{c}{\textbf{Average}} \\ 
Cascade Grid & \textbf{54.29} & 55.18&25.93 & 45.13 \\
Complete Cover & 49.04	& \textbf{55.79}	& \textbf{28.21} & 44.34  \\
ViT & 54.16 & 55.09 & 26.36 & \textbf{45.20} \\
\bottomrule
\end{tabular}%
\end{table}

Table \ref{table:patchdiv} presents the performance of different patch division algorithms. "Complete Cover" achieves the highest average Counting and Color score, primarily attributed to its diverse patch scales. "Cascade Grid" offers the best average Position accuracy, likely due to its effective balance between patch scales and the number of distractor patches (better than dense "Complete Cover"). 
ViT demonstrates the most balanced performance, likely due to the small number of patches reducing potential distractions.

\subsection{Proxy Selection Function}
\begin{table}[htbp]
\caption{Regular VQA Task Result with different proxy selection functions.}
\label{table:proxy}
\centering
\begin{tabular}{ccccc}
\toprule \Gray
 \multicolumn{1}{c}{\textbf{Proxy Selection Function}} & \multicolumn{4}{c}{\textbf{Accuracy}} \\ \Gray
  \multicolumn{1}{c}{Task Split} & \multicolumn{1}{c}{\textbf{Position}} & \multicolumn{1}{c}{\textbf{Color}} & \multicolumn{1}{c}{\textbf{Count}} & \multicolumn{1}{c}{\textbf{Average}} \\ 
Prototype & 52.12 & 55.16 & \textbf{27.58} & \textbf{44.95}\\
Reranking & \textbf{52.47} & 55.13 & 26.95 & 44.85 \\
Clustering &  51.10 & \textbf{56.19} & 26.28 & 44.52  \\
\bottomrule
\end{tabular}%
\end{table}

Table \ref{table:proxy} shows the performance of different proxy feature selection functions. Overall, their performance is comparable, yet each excels in specific tasks: "Reranking" in Position, 
"Clustering" in Color, and "Prototype" in Counting. In practice, high-accuracy trails mostly come from "Prototype" and "Clustering". However, since "Clustering" is much more time-consuming, "Prototype" is more recommended.

\subsection{ROI size meets MLLM with visual cues}
\label{sec:roi_size}
\begin{figure}[htbp]
    \centering
    \includegraphics[width=0.45\textwidth]{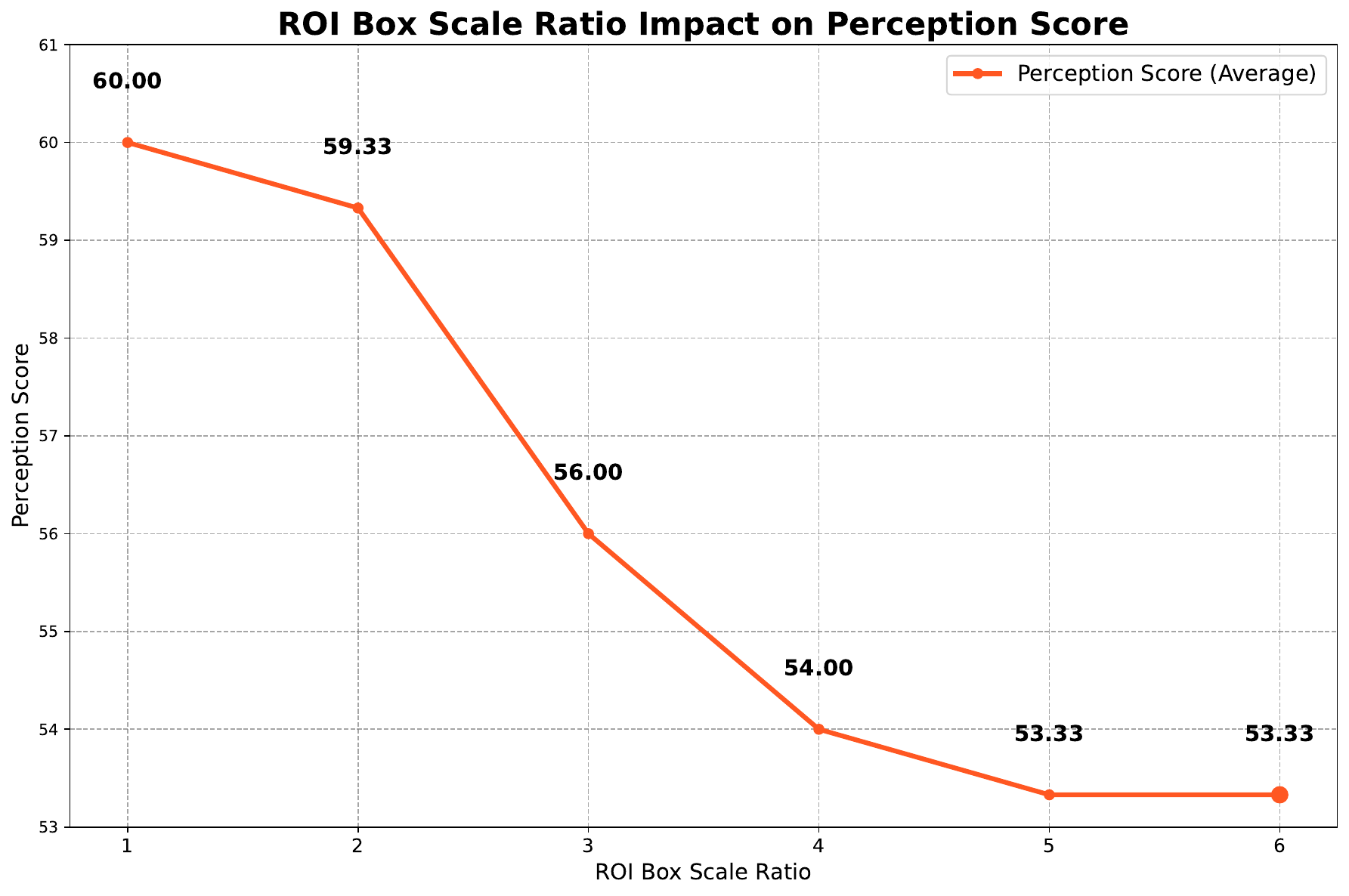}
    \caption{A demonstration of \textbf{Zoom4K} and how to fine-tune Multimodal Large Language Model with visual cues}
    \label{fig:roi_enlarge}
\end{figure}

To study how enlarging the ROI size affects the inferring model's performance, we expanded the width and height of the ROI box using multipliers from 1 to 6, significantly increasing the ROI area while keeping its center unchanged. As shown in Figure \ref{fig:roi_enlarge}, the overall accuracy of the regular VQA task declined substantially with larger ROI sizes. This is because the expanded visual cues become less precise and lose critical details, making them less effective for the inferring model. This experiment demonstrates that we shouldn't provide the inferring model with overly large visual cues, even though they contain the ground truth ROI box.

\section{Result Visualization and Discussion}

\begin{figure*}[htbp!]
    \centering
    \includegraphics[width=0.9\textwidth]{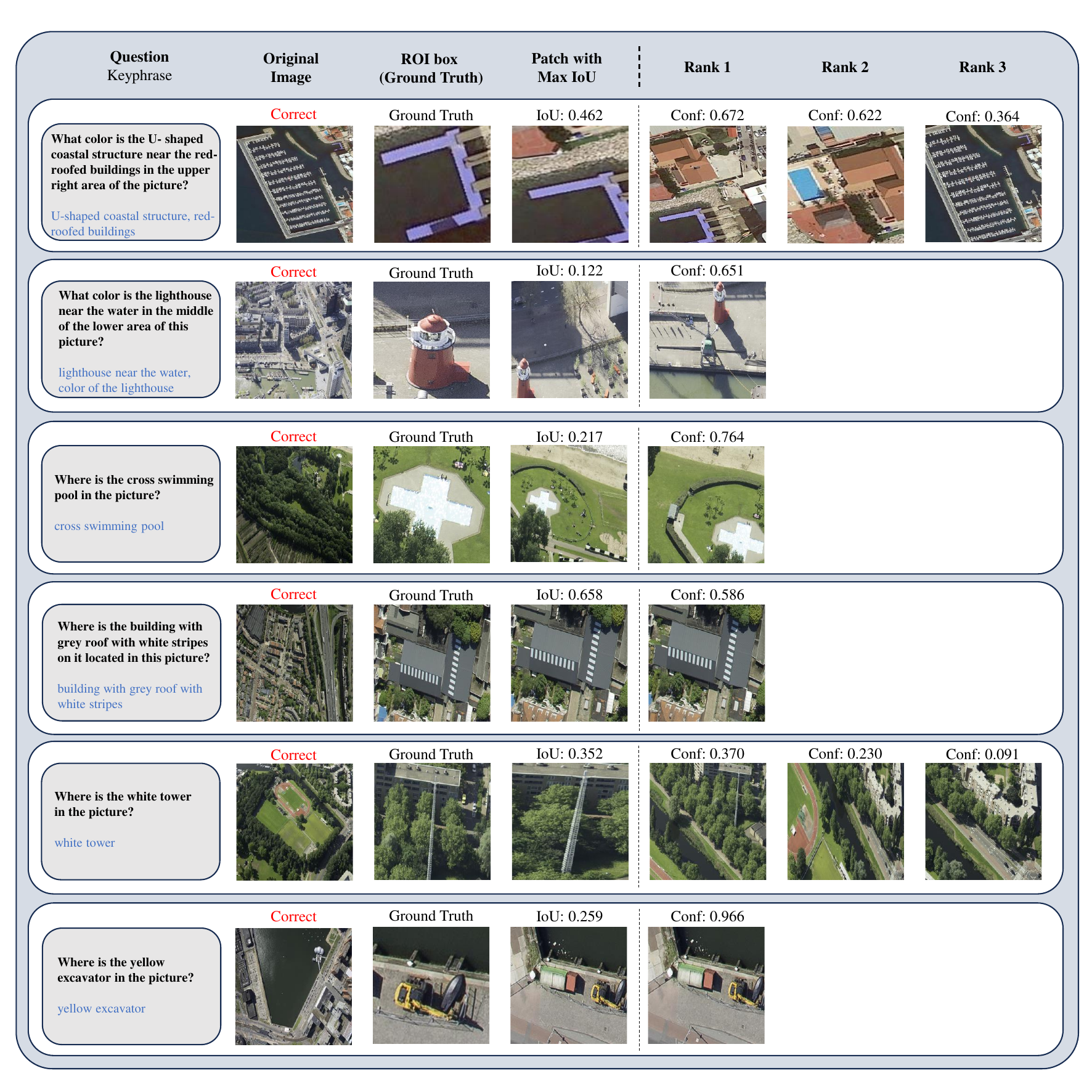}
    \caption{Good retrieval results with correct responses.}
    \label{fig:goodgood}
\end{figure*}

\begin{figure*}[htbp!]
    \centering
    \includegraphics[width=0.92\textwidth]{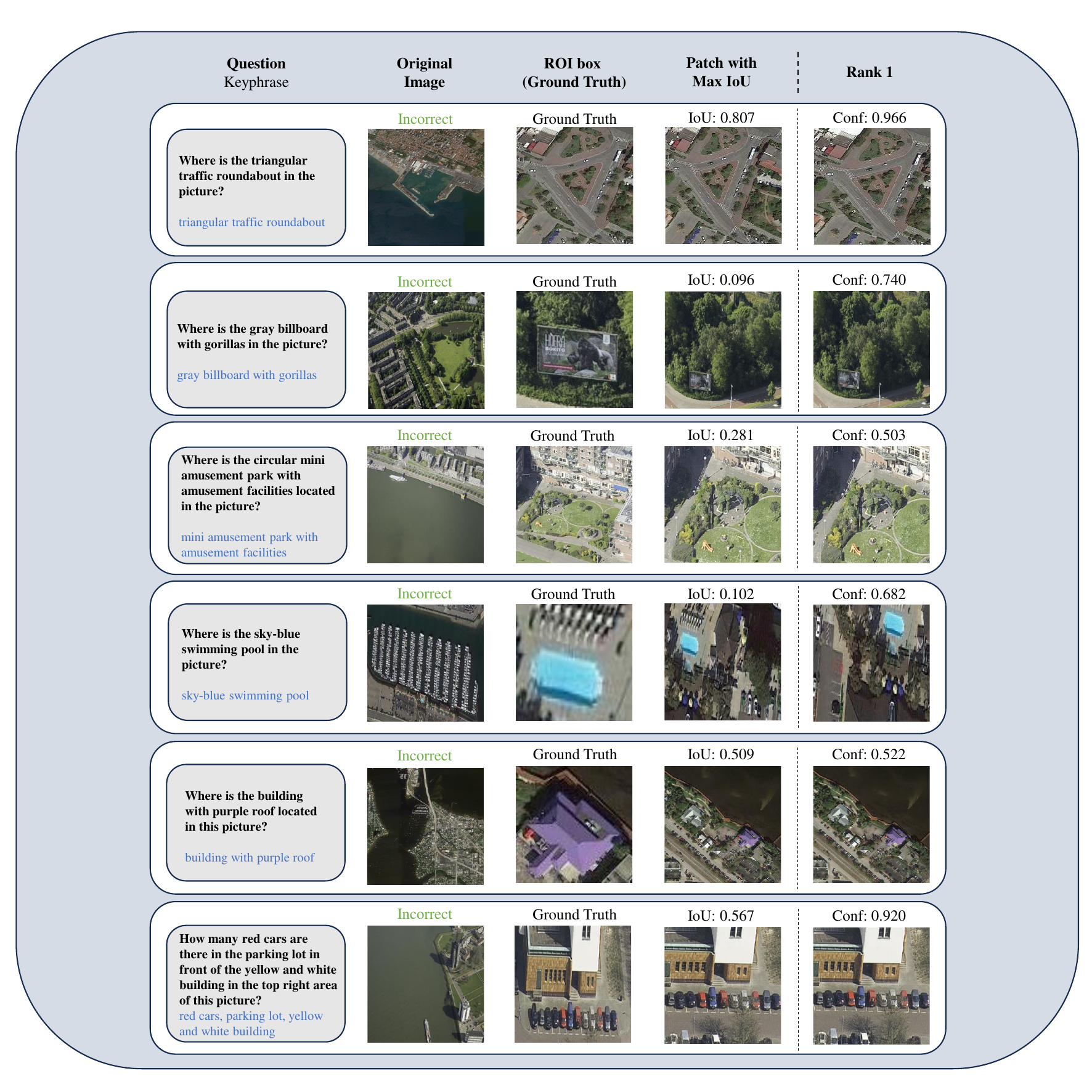}
    \caption{Good retrieval results with incorrect responses.}
    \label{fig:goodbad}
\end{figure*}

In Figures \ref{fig:goodgood} and \ref{fig:goodbad}, we visualize correct and incorrect responses with high-quality retrieval results. For each QA pair, we list the questions and key phrases from the Question Analyzing Module. We also include the ground truth ROI box and the image patch with the maximum IoU in relation to the ROI box. For each question, the top-3 retrieved visual cues are displayed alongside their confidence scores.

From both figures, we can see that the Question Analyzing Module does a good job of extracting key phrases from questions. The Image-Text retrieval module and Image-Image retrieval module can obtain useful visual cues, some of which even overlap with the best available patch (the one with the largest IoU with the ROI box among all patches). The ImageRAG framework can identify rare concepts in remote sensing, such as lighthouses, excavators, coastal structures, even \textbf{Gorillas}. All the retrieved objects are extremely small in the original UHR RSI.

In Figure \ref{fig:goodgood}, the rank 1 retrieved results are sufficient for directly answering the Color and Counting questions. The inferring model can interpret the positional relationship between local and global images when coordinates are provided. The visual connection of the visual cues to the question aligns with their confidence scores.

Figure \ref{fig:goodbad} is particularly interesting. In the case of the triangular roundabout, the target can be located precisely, with the patch fitting tightly with the ROI box. However, despite this accurate visual cue, the inferring model fails to answer correctly, suggesting room for improvement in its positional inferring capabilities. Regarding the Gorillas example, ImageRAG retrieves the correct patch, though not perfectly aligned and slightly zoomed out. Yet, the model cannot overcome the trap in positional descriptions (the correct answer is "At the top of the Y-shaped intersection," which actually refers to the bottom part of this Y-shaped intersection). Moreover, even with a perfect patch, the framework doesn't further enhance the inferring model's counting ability, as it fails to count "red cars" although the best patch is found.

\section{Computational Resource}
The LoRA fine-tuning process took 2 hours for 2 epochs using 8 NVIDIA A100-80GB GPUs \footnote{https://internvl.readthedocs.io/en/latest/internvl2.5/finetune.html}, and hosting Qwen-32B-Instruct required 4 NVIDIA A100-40GB GPUs. As for running ImageRAG, a single NVIDIA A100-40GB or NVIDIA RTX 4090-24GB would be enough.

\section{A Cookbook for Adapting ImageRAG to Different Image Modalities and  Domains}
\textcolor{black}{
In this paper, we focus on the optical ultra high resolution remote sensing
image, but ImageRAG is a geneal framework that is able to be applied in different UHR image modalities (e.g. SAR and Hyperspectral data), and image domains (e.g. Medical Imaging). Several modules can be redesigned to better fit the specific task, and some others can remain invariant.
}

\subsection{Patchify}
\textcolor{black}{
In section \ref{sec:pd}, we introduced three selected patch division algorithm: "ViT", "Complete Cover", "Cascade Grid". These are model-agnostic approaches, which means it can be adopted to different image modalities and domains without any modification or retraining. 
}

\subsection{Question Analyzing Module}
\label{sec:cookbook_qam}
\textcolor{black}{
We deploy Qwen2.5-32B on SGLang for API serving to extract the target-of-interest from questions, ensuring the extracted key phrases are as accurate as possible. However, since the target-of-interest extraction task is relatively simple, this model could potentially be replaced by lighter general LLMs such as MiniCPM3-4B \cite{hu2024minicpm} and Qwen3-0.6B in real-world deployment scenarios, or domain-specific LLM like Med-PaLM 2 \cite{Singhal2025-vh} when dealing with UHR medical images.
}

\subsection{Text-Image Retrieval Module}
\textcolor{black}{
This module is of critical importance to the ImageRAG framework, since image patches, target-of-interest text, image data from vector database are all encoded as feature vectors using the VLM within this module. More importantly, the visual cues are retrieved based on the similarities between these features. In our study, we tried domain-specific contrastive VLMs (RemoteCLIP, GeoRSCLIP) and general VLMs (CLIP, MCIPCLIP) for UHR optical remote sensing images. For encoding SAR, Multispectral, infrared or hyperspectral imaging alongside corresponding text, GeoLangBind \cite{xiong2025geolangbindunifyingearthobservation} can be considered. As for medical imaging from other image domain, MedCLIP \cite{wang2022medclipcontrastivelearningunpaired} can be chosen. 
}

\subsection{Text-Text Retrieval Module and Vector Database}
\textcolor{black}{
This module matches the data from the vector database with the key phrases extracted from the question. Sentence-BERT is generally suitable for text-text retrieval. When dealing with modality-specific or domain-specific data, it is advisable to collect images accompanied by text labels or captions to construct the vector database effectively. 
}
\textcolor{black}{
In scenarios where access the image-text paired data is limited, the calculation of proxy image embeddings can be replaced by leveraging all available data through in-context few-shot learning \cite{flamingo}, which involves constructing task examples alongside the question to send to the VLM. This approach aligns with the ultimate goal of the slow path, which is to select useful image patches with the aid of external data as visual cues to send to the MLLM.
}



\section{Limitation}
\textcolor{black}{
The ImageRAG framework has two major limitations. First, the time cost is significant. Second, the performance heavily depends on the accuracy of retrieved visual cues. 
}

\begin{table}[htbp]
\caption{We compared the time cost of ImageRAG across different strategies. "Mode" shows whether ImageRAG used only the fast path or a combination of fast and slow paths as detailed in Section \ref{sec:retrievalstage}. "Vector Database" reflects if LRSD and CRSD were used as described in Section \ref{sec:ttrm}. "Time Cost" represents the average processing time per question in seconds. All patches and their corresponding features are pre-extracted and used as caches. Vector Databases are pre-built.
}
\label{table:computation_time}
\centering
\begin{tabular}{cccc}
\toprule \Gray
 \multicolumn{1}{c}{\textbf{Mode}} & \multicolumn{1}{c}{\textbf{Vector Database}} & \multicolumn{1}{c}{\textbf{Time Cost }} & \multicolumn{1}{c}{\textbf{Note}}\\ 
Baseline & - & 1.14 sec & Direct Inference \\
Fast Path Only & - & 2.70 sec & All Fast Path\\
Fast + Slow Path & LRSD & 2.89 sec  & All Slow Path\\
Fast + Slow Path & LRSD + CRSD & 3.02 sec & All Slow Path\\
Fast + Slow Path & LRSD + CRSD & 2.81 sec & Fast \& Slow Path \\

\bottomrule
\end{tabular}%
\end{table}

\textcolor{black}{
As shown in Table \ref{table:computation_time}, we compared the relative time cost of ImageRAG across different strategies. For fair comparison, we pre-extracted all patches and their corresponding features and used them as caches (This process is parallelized using Ray \footnote{https://github.com/ray-project/ray} with multiple GPUs). Vector Databases are pre-built. Compared with the baseline setting, which does not use ImageRAG, the additional time cost primarily comes from question analyzing and visual cue retrieval. 
When using the fast path for all questions, the time cost doubles (1.14 seconds compared to 2.70 seconds). If all questions require the worst-case scenario—where the fast path fails and the slow path must search both vector databases to find visual cues—the time cost is almost tripled (1.14 seconds compared to 3.02 seconds). In practical scenarios, when mixing fast and slow paths across the dataset, the time cost is nearly 2.5 times higher than the baseline (1.14 seconds compared to 2.81 seconds). 
However, the absolute time cost should be acceptable in practical development as the analysis processing time remains measured in seconds. To further enhance speed, the Qwen2.5-32B model used for question processing can be replaced with a lighter model as mentioned in section \ref{sec:cookbook_qam}.
}

\textcolor{black}{
Moreover, as shown in Section \ref{sec:infer_vqa} and Section \ref{sec:roi_size}, the performance of ImageRAG is significantly affected by the accuracy and size of the retrieved visual cues. When the retrieved visual cues are overly large or  incorrect, the performance of ImageRAG can greatly decline (Figure \ref{fig:roi_enlarge}) and may even drop below the baseline (as seen in the "Random Box" row of Table \ref{tab:inferringvqa}). Techniques referenced in section \ref{sec:relaworl} that improved the accuracy of the retrieved results of RAG can also be applied to mitigate this issue.
}

\section{Related Work}

\subsection{Remote Sensing meets Vision-Language Models}

Remote Sensing Vision-Language Models (RSVLMs) are developed to analyze geospatial data by incorporating both visual and linguistic information~\cite{li2024visionlanguagemodelsremotesensing}. These models are pre-trained on large-scale RSI and relevant text, enabling them to adapt to various VLM tasks in RS domains. RSVLMS have demonstrated promising results in specialized applications, including scene classification~\cite{earthgpt, remoteclip}, object detection~\cite{geochat, skysensegpt}, semantic segmentation~\cite{yuan2024rrsisreferringremotesensing, liu2024rotatedmultiscaleinteractionnetwork}, image captioning~\cite{earthgpt, geochat}, text-image retrieval \cite{rs5m, remoteclip}, visual grounding~\cite{geochat, vrsbench, skysensegpt}, image generation~\cite{rs5m, khanna2024diffusionsatgenerativefoundationmodel, yu2024metaearthgenerativefoundationmodel}, etc. Many of them are capable to complete multiple downstream tasks.

RSVLMs can be categorized into three types based on their input-output mechanisms ~\cite{zhou2024visionlanguagegeofoundationmodelsurvey}. Contrastive RSVLMs process both text and images as inputs, generating similarity scores essential for tasks such as image-text retrieval and zero-shot scene classification. Conversational (generative) RSVLMs also take text and images as inputs but produce textual responses, leveraging LLMs for tasks like captioning and visual question answering. Besides, some RSVLMs are conditioned on either text or images to generate synthetic remote sensing images, typically employing conditional diffusion processes for controlled image synthesis.


For contrastive RSVLMs, research primarily focuses on extending CLIP~\cite{clip} to RS applications, emphasizing the development of RS-specific datasets and benchmarks. Liu et al. proposed RemoteCLIP~\cite{remoteclip}, a vision-language foundation model for remote sensing that achieves significant improvements in various downstream tasks through multi-task pre-training and data expansion. Zhang et al. developed GeoRSCLIP~\cite{rs5m} based on the large-scale remote sensing image-text dataset RS5M, which demonstrates excellent performance in zero-shot classification, cross-modal text-image retrieval, and semantic localization tasks. Mall et al. introduced GRAFT~\cite{graft}, a method to train vision-language models for satellite images without textual annotations by using ground images as an intermediary. Wang et al. constructed SkyScript~\cite{Wang_Prabha_Huang_Wu_Rajagopal_2024}, a large and semantically diverse vision-language dataset for remote sensing, and developed SkyCLIP through continual pre-training, which shows superior performance in zero-shot scene classification, fine-grained attribute classification, and cross-modal retrieval tasks.

A typical conversational RSVLM comprises three main components: a pre-trained visual encoder, a pre-trained LLM, and a modality interface that connects them. RSGPT~\cite{hu2023rsgptremotesensingvision} constructs a high-quality remote sensing vision-language model by leveraging large-scale image-text pairs for pre-training, focusing on image captioning and visual question answering tasks. GeoChat~\cite{geochat} develops a novel multimodal framework that unifies various remote sensing tasks through visual perception and language model alignment, achieving state-of-the-art performance on multiple benchmarks. SkyEyeGPT~\cite{ZHAN202564} proposes a unified multimodal framework that leverages instruction tuning to handle diverse remote sensing tasks, showing superior performance across different granularity levels. EarthGPT~\cite{earthgpt} creates a versatile multimodal large language model tailored for remote sensing, integrating multi-sensor data and multi-task instruction tuning to excel in tasks. LHRS-Bot~\cite{10.1007/978-3-031-72904-1_26} builds large-scale remote sensing datasets and employs a multi-level vision-language alignment strategy to enhance model performance, establishing a comprehensive benchmark for evaluating multimodal models in the remote sensing domain. RS-CapRet~\cite{silva2024largelanguagemodelscaptioning} employs a frozen LLM and a contrastively trained vision encoder for efficient fine-tuning in remote sensing image captioning and retrieval. VHM~\cite{pang2024vhmversatilehonestvision} enhances remote sensing analysis with rich-captioned and honest-question datasets, improving factual consistency in vision-language tasks. SkySenseGPT~\cite{skysensegpt} is optimized for fine-grained relation understanding, leveraging the large-scale FIT-RS instruction dataset.

Generative foundation models, particularly diffusion models, constitute a fundamental class in image generation tasks and offer significant potential for advancing remote sensing applications. DiffusionSat~\cite{khanna2024diffusionsatgenerativefoundationmodel} presents a large-scale generative foundation model for satellite imagery, integrating text and metadata conditioning to enable high-resolution generation, super-resolution, temporal generation, and inpainting, outperforming previous models. CRS-Diff~\cite{tang2024crsdiffcontrollableremotesensing} introduces a controllable remote sensing image generation model based on diffusion models, incorporating multi-modal conditioning to achieve precise, high-quality, and flexible RS image synthesis. Changen2~\cite{10713915} proposes a generative change foundation model for remote sensing change detection, leveraging a probabilistic graphical model and diffusion transformers to synthesize realistic multi-temporal change data, enabling zero-shot change detection and improved transferability.

\subsection{Vision-Language Model with Visual Cues}

In our default setting, we adopt the approach for the VQA LLM similar to $V^*$~\cite{wu2024v} and Zoom Eye~\cite{shen2024zoomeye}, emphasizing the perception of visual cues by incorporating their sub-patches after considering the global view of the entire image.

Apart from this, a cascade of alternative methods could guide the attention of MLLMs to specific regions within an image. For instance, the VQA LLM could be substituted with training-free models like ControlMLLM~\cite{wu2024controlmllmtrainingfreevisualprompt}, which assigns attention scores to a particular region of the image, thereby directing the model’s focus to those localized areas. In addition, methods that integrate coordinates of bounding boxes, points, or even visual features from local regions as supplementary visual cues may be employed to enhance the model’s regional awareness. Notable examples of such approaches include Kosmos-2~\cite{peng2023kosmos}, Shikra~\cite{chen2023shikra}, and Ferretv2~\cite{zhang2024ferretv2improvedbaselinereferring}. 

Furthermore, conventional MLLMs typically rely on  patch-level vision encoders to encode the input image~(e.g., Vision Transformer~\cite{dosovitskiy2020image}), which may perform suboptimally in regional information processing. Inspired by related works on region-based vision-language pre-training~\cite{chen2020uniter,zhang2021vinvl} that achieves superior location understanding in virtue of a pre-trained object detector~\cite{ren2016faster}, we could incorporate visual features extracted from the recent outstanding object detectors\cite{zhou2022detecting, liu2023grounding} into our VQA LLM to enhance it with object-enteric visual cues understanding. A similar approach has been explored in ChatRex~\cite{jiang2024chatrextamingmultimodalllm}.

\subsection{Retrieval-Augmented Generation}
\label{sec:relaworl}
Retrieval-Augmented Generation (RAG) addresses the limitations of traditional generative models in handling specialized or long-tail knowledge. Early models like GPT, trained on vast corpora, excel at general queries but struggle with domain-specific or rare information, often generating hallucinations~\cite{zhang2023siren}. RAG, introduced by Facebook AI Research in 2020~\cite{lewis2020retrieval}, enhances generative models by integrating real-time document retrieval, improving accuracy and contextual grounding. Gao et al.\cite{gao2023retrieval} categorize RAG into Naive, Advanced, and Modular paradigms, detailing key components like retrievers, generators, and augmentation methods. A comparative study by Ovadia et al.\cite{ovadia2023fine} shows that RAG outperforms unsupervised fine-tuning, particularly in scenarios involving new or unseen knowledge, underscoring its superiority in knowledge injection and model adaptation.


The effectiveness of RAG systems heavily depends on the quality and relevance of the retrieved knowledge, which directly influences the accuracy and factual grounding of generated content. To enhance retrieval efficiency and overcome the limitations of traditional methods, several advancements have been proposed, particularly for zero-shot and few-shot retrieval tasks. Techniques such as HyDE~\cite{gao2023precise} and REINA~\cite{wang2022training} utilize LLMs to generate hypothetical documents, improving retrieval performance without requiring labeled data. The Rewrite-Retrieve-Read ~\cite{ma2023query} framework introduces a query rewriting step, allowing the input query to be better aligned with retrieval modules. By using reinforcement learning to adapt queries, R3 enhances retrieval quality, improving performance in open-domain and multiple-choice question answering tasks. Promptagator~\cite{dai2022promptagator} demonstrates the effectiveness of few-shot learning in dense retrieval, utilizing LLMs to generate synthetic training data from minimal examples, surpassing models trained on large-scale datasets like MS MARCO. This underscores the viability of few-shot learning and LLM-generated synthetic data in resource-constrained settings. To bridge the preference gap between retrievers and LLMs, Zixuan Ke et al.~\cite{ke2024bridging} introduce the BGM framework, which employs a sequence-to-sequence model to align retrieved information with LLM preferences. 
 
Methods like RECITE~\cite{sun2022recitation} and ITER-RETGEN~\cite{shao2023enhancing} focus on improving factual accuracy and integrating retrieved knowledge, ensuring grounded, accurate responses through knowledge recitation and iterative retrieval-generation. Frameworks such as Selfmem~\cite{cheng2023lift} and Self-RAG~\cite{asai2023self} enhance factual consistency via self-reflection. Techniques like Step-Back Prompting~\cite{zheng2023take} improve reasoning by guiding LLMs to abstract concepts. The GENREAD approach~\cite{yu2022generate} replaces traditional retrieval with LLM-generated contextual documents, demonstrating superior performance in knowledge-intensive tasks like open-domain QA and fact-checking.

Iterative and active retrieval-generation strategies aim to dynamically enhance both retrieval and generation processes. ITER-RETGEN~\cite{shao2023enhancing} alternates between retrieval and generation to improve response quality, while FLARE~\cite{jiang2023active} predicts and adapts to future information needs during generation. These methods boost relevance and accuracy in knowledge-intensive tasks. The Adaptive-RAG framework\cite{jeong2024adaptive} selects retrieval strategies based on query complexity, outperforming static models in efficiency and accuracy. 
FunnelRAG \cite{zhao-etal-2025-funnelrag} proposes a progressive retrieval paradigm with coarse-to-fine granularity for RAG, to enable load balancing and improve retrieval performance.

Reducing hallucinations in response is a critical challenge in RAG systems. To address this, various strategies focus on improving the trustworthiness of generated outputs by leveraging more reliable retrieval mechanisms and robust post-generation processes. Methods such as RAGTruth~\cite{niu2023ragtruth} introduce large-scale datasets designed to detect and mitigate hallucinations, enabling models to generate more trustworthy responses. SEER \cite{zhao-etal-2024-seer} proposes a novel evidence extraction learning paradigm, which utilizes the model to calibrate its extraction preference via self-alignment. Ayala et al. ~\cite{bechard2024reducing} use external knowledge sources to reduce errors in structured output generation, enhancing the reliability of RAG systems in practical applications. 


Task-specific advancements in RAG systems focus on refining retrieval-augmented models for particular applications, improving their efficiency and effectiveness in complex tasks. Demonstrate-Search-Predict~\cite{khattab2022demonstrate} introduces a modular approach that breaks down complex problems into manageable tasks, enhancing performance in multi-hop reasoning and open-domain question answering. Similarly, RA-DIT~\cite{lin2023ra} uses dual instruction tuning to fine-tune the retriever and generative model, optimizing their collaboration for knowledge-intensive benchmarks. These methods highlight the importance of tailoring RAG systems to specific tasks, enabling more effective and accurate solutions across diverse domains.



\subsection{Multimodal RAG}

Multimodal RAG technology is an extension of the traditional RAG model, designed to enhance the performance of generative tasks by incorporating information from multiple data modalities~\cite{abootorabi2025ask}. Unlike the traditional RAG, which processes only textual data, multimodal RAG can handle not only text but also other modalities such as images, audio, and video. It is capable of extracting information from various modalities and integrating this information to generate richer and more accurate outputs. In multimodal RAG systems, embeddings for various data types, such as text and images, are generated through modality-specific encoders~\cite{mortaheb2025re}. These encoders share a unified embedding space, which is also employed for encoding the query.

The latest advancements in RAG in the image domain have led to significant improvements ~\cite{zheng2025retrievalaugmentedgenerationunderstanding}. RA-CM3~\cite{yasunaga2022retrieval} enhances both text-to-image and image-to-text generation by combining the CLIP retriever and the CM3 Transformer generator, achieving a performance boost while reducing computational costs by over 30\%. Mortaheb et al. introduced a re-ranking mechanism based on a relevance score model~\cite{mortaheb2025rag}, improving context selection during retrieval and reducing hallucinations, thereby enhancing the quality of generated responses. Yu et al.'s VisRAG~\cite{yu2024visrag} framework bypasses the text parsing stage to directly process multi-modal documents containing both text and images, achieving substantial improvements in multi-modal tasks. Bonomo and Bianco's Visual RAG~\cite{bonomo2025visual} expands the visual knowledge of large MLLMs without the need for fine-tuning by dynamically retrieving relevant examples, offering high computational efficiency. Riedler and Langer's work on multimodal inputs for industrial applications demonstrates that integrating both images and text in RAG systems significantly improves performance~\cite{riedler2024beyond}.

RAG has also made significant progress in the video domain, driving innovations in long video comprehension. Yongdong Luo et al. propose Video-RAG~\cite{luo2024video}, which enhances long video understanding by integrating visually-aligned auxiliary texts into large video-language models, surpassing models like Gemini1.5-Pro and GPT-4o. Jeong et al. introduce VideoRAG\cite{jeong2025videorag}, a method that dynamically retrieves relevant videos based on user queries and combines both visual and textual information to generate more contextually rich responses, showing marked improvements over traditional RAG approaches. Ma et al. present DrVideo\cite{ma2024drvideo}, a system that converts long videos into text-based documents and iteratively retrieves missing information, achieving high accuracy in key frame identification. Arefeen et al. introduce iRAG\cite{arefeen2024irag}, an incremental method that accelerates video-to-text processing by using lightweight models for fast indexing and heavyweight models for detailed extraction, making it well-suited for real-time video analysis. Zhang et al. propose OmAgent\cite{zhang2024omagent}, a multi-modal agent framework that minimizes information loss in video understanding tasks by dynamically invoking retrieval tools, thereby enhancing reasoning and event localization.

\section{Conclusion and Future Work}

In this work, we introduced the \textbf{ImageRAG} framework. It retrieves relevant visual context from UHR remote sensing images based on key phrases in text queries, enabling the MLLM to focus on \textbf{important details, including tiny ones}, and answer questions by inferring them. ImageRAG integrates various \textbf{external knowledge databases} to guide the model, enhancing its understanding of the query and UHR RSI. Notably, ImageRAG requires minimal training (only fine-tuning the inferring model), making it a practical solution for efficiently handling UHR RSI. We also introduce the MME-RealWorld-lite-RS benchmark, demonstrating that ImageRAG achieves strong performance across Regular VQA, Inferring VQA, and Visual Cue Retrieval tasks on it.

In the future, we plan to apply more RAG techniques to ImageRAG, particularly enhancing the ranking component. We will also prioritize optimizing ImageRAG's efficiency and scalability. For the inferring model, we will investigate how to minimize the negative influence of inaccurate visual cues. Moreover, to boost performance on specialized imaging data, we intend to introduce adapter module with trainable parameters. This will enable effective adaptation to domain-specific data such as SAR, infrared, and hyperspectral imagery. The approach may adopt an in-context learning style, minimizing the need for extensive additional training data while preserving the model's general capabilities.

\clearpage
\appendices
\section{MME-RealWorld-Lite-RS Corrections}
\label{sec:appendix_correction}
Two majority corrections are "Multiple Correct Answers" and "Incorrect Label".
\begin{figure}[h]
\begin{tcolorbox}[colback=gray!10, colframe=black!50, width=0.5 \textwidth, sharp corners]
\textbf{Multiple Correct Answers:} \par

\texttt{Question\_id: \\
    perception/remote\_sensing/color/3571} \par
    Original: "B" \par
    Corrected: ["B", "D"] \par

\texttt{Question\_id: \\
    perception/remote\_sensing/color/2474} \par
    Original: "A" \par
    Corrected: ["A", "B"] \par
\end{tcolorbox}
\end{figure}

\begin{figure}[h]
\centering
\begin{tcolorbox}[colback=gray!10, colframe=black!50, width=0.5\textwidth, sharp corners]
\textbf{Incorrect Label:} \par

\texttt{Question\_id: \\
    perception/remote\_sensing/position/3639} \par
    Original: "(B) In the right left area of the picture" \par
    Corrected: "(B) In the upper left area of the picture" \par

\texttt{Question\_id: \\
    perception/remote\_sensing/position/3270} \par
    Original: "(B) In the lower right corner of the picture" \par
    Corrected: "(B) In the lower left area of the picture" \par

\texttt{Question\_id: \\
    perception/remote\_sensing/color/0541} \par
    Original: "(A) Green" \par
    Corrected: "(A) White" \par
\end{tcolorbox}
\end{figure}

\section{MME-RealWorld Prompt Template}
\label{appendix:pt_mmerealworld}
\begin{figure}[htbp!]
\begin{tcolorbox}[colback=gray!10, colframe=black!50, width=0.5\textwidth, sharp corners]
\textbf{Question Prompt:} \par
Select the best answer to the above multiple-choice question based on the image. Respond with only the letter (A, B, C, D, or E) of the correct option. \par
\{\textbf{options}\} \par
\textbf{Chain-of-Thought Question Prompt:} \par
Select the best answer to the above multiple-choice question based on the image. \textit{Provide the thinking process and give the response in the end} with the letter (A, B, C, D, or E) of the correct option." \par
\{\textbf{options}\} \par
\end{tcolorbox}
\end{figure}

\section{Inferring Model Prompt Template}
\label{appendix:pt_inferringmodel}
\begin{figure}[htbp!]
\begin{tcolorbox}[colback=gray!10, colframe=black!50, width=0.49\textwidth, sharp corners]
\textbf{Prompt Template for Inferring Model:} \par
<image> \par
Additional information: \par
Sub-patch 1 at location <box>[[x1, y1, x2, y2]]</box>: <image> \par
Sub-patch 2 at location <box>[[x3, y3, x4, y4]]</box>: <image> \par
Sub-patch 3 at location <box>[[x5, y5, x6, y6]]</box>: <image> \par
Look at the image and answer the question based on the provided additional information (location of sub-patches). \par
Question: \par
\{\textbf{question with options}\} \par
\end{tcolorbox}
\end{figure}

\newpage
\section{Prompt Template for Question Analyzing Module}
\label{sec:appendix_prompt_qam}
\begin{figure}[htbp!]
\begin{tcolorbox}[colback=gray!10, colframe=black!50, width=0.48\textwidth, sharp corners]
\textbf{Prompt Template for Keyphrase Extraction:}

\noindent
Task: Extract keywords or key phrases from a given query sentence.

Guidelines: \\
- Analyze the sentence and identify the important keywords or phrases. These words or phrases should represent the core content or main information of the sentence. \\
- Ensure that the extracted words or phrases are meaningful. Focus on the names of ground targets from a remote sensing perspective. \\
- All adjectives should be part of a phrase that includes a target.\\
- Do not include standalone adjectives such as adjectives that describe size, shape, color, texture, etc. unless they are paired with a target. \\
- Do not include keywords or phrases that relate to position and orientation.\\
- Avoid including overly vague words such as 'image', 'picture', 'photo', 'object', etc., unless they are part of a more specific phrase that provides additional context.

Example:\\
Input query:  "How many aircraft are heading in the up-right direction? What is the color of the building rooftop below them?"\\
Output: ["aircraft", "building rooftop", "aircraft are heading in the up-right direction", "color of the building rooftop"]

Your task is to apply the same process to the following sentence:

Input query: \{\textbf{question}\}\\
Output:
\end{tcolorbox}
\end{figure}

\newpage
\section{Class Name in LRSD}
\label{sec:appendix_LRSD}
\begin{figure}[htbp!]
\begin{tcolorbox}[colback=gray!10, colframe=black!50, width=0.48\textwidth, sharp corners] 
\textbf{Category List:} \par
\vspace{0.2cm} 
\noindent road, parking lot, tugboat, nuclear powerplant, a220, meadow, police station, gas station, engineering ship, lighthouse, ignore, multi unit residential, boeing737, quarry, container, airport terminal, wind farm, zoo, roundabout, bridge, debris or rubble, apron, substation, race track, shipyard, swimming pool, car dealership, water, truck tractor, airplane, forest, soccer ball field, solar farm, a330, windmill, airport hangar, tennis court, plane, park, viaduct, surface mine, container crane, other ship, crop field, ground track field, barn, cargo truck, bus, a321, large vehicle, warship, dry field, train station, c919, boeing787, helipad, church, baseball diamond, space facility, oil field, dump truck, amusement park, office building, electric substation, van, other vehicle, archaeological site, fountain, smokestack, dry cargo ship, excavator, pier, place of worship, wind turbine, railway bridge, stadium, small vehicle, recreational facility, water treatment facility, interchange, helicopter, liquid cargo ship, factory or powerplant, aquaculture, construction site, toll booth, boeing747, border checkpoint, passenger ship, prison, works, airport, agriculture, fishing boat, hospital, runway, ship, storage tank, burial site, motorboat, orchard, waste disposal, tower, flooded road, arj21, background, other airplane, small car, mine, boeing777, shopping mall, oil or gas facility, terraced field, solar power plant, educational institution, single unit residential, intersection, tunnel opening, road bridge, building, wastewater plant, basketball court, ground transportation station, tractor, military facility, railway, dam, fire station, greenhouse, parking lot or garage, harbor, baseball field, port, barren, paddy field, cemetery, trailer, a350, golf course, impoverished settlement, lake or pond, football field.
\end{tcolorbox}
\end{figure}

\clearpage

\bibliographystyle{IEEEtran}
\bibliography{ref}

\end{document}